\documentclass[10pt,journal,compsoc]{IEEEtran}
\ifCLASSINFOpdf
\else
\fi

\usepackage[square,numbers]{natbib}
\usepackage{hyperref}
\usepackage{amssymb} 
\usepackage{mathalfa} 
\usepackage{algorithm} 
\usepackage{algpseudocode} 
\usepackage[table]{xcolor}
\usepackage{array}
\usepackage{titlesec}
\usepackage{mdframed}
\usepackage{listings}
\usepackage{comment}
\usepackage{amsmath} 
\usepackage{booktabs}
\usepackage{multirow}
\usepackage{tabularx}
\usepackage{tabularray}
\UseTblrLibrary{diagbox}
\usepackage{dsfont}
\usepackage{array}
\usepackage[capitalise]{cleveref}
\usepackage{graphicx}
\usepackage{float}
\usepackage{subfig}
\usepackage{lscape}

\usepackage{amssymb}

\newcolumntype{L}[1]{>{\raggedright\let\newline\\\arraybackslash\hspace{0pt}}m{#1}}
\newcolumntype{C}[1]{>{\centering\let\newline\\\arraybackslash\hspace{0pt}}m{#1}}
\newcolumntype{R}[1]{>{\raggedleft\let\newline\\\arraybackslash\hspace{0pt}}m{#1}}

\usepackage[nolist,nohyperlinks]{acronym}

\begin{document}

\title{A Survey on the Robustness of Computer Vision Models against Common Corruptions}

\author{Shunxin~Wang,
        Raymond~Veldhuis,
        Christoph~Brune,
        Nicola~Strisciuglio}

\maketitle

\begin{abstract}

The performance of computer vision models are susceptible to unexpected changes in input images caused by sensor errors or extreme imaging environments, known as common corruptions (e.g. noise, blur, illumination changes). These corruptions can significantly hinder the reliability of these models when deployed in real-world scenarios, yet they are often overlooked when testing model generalization and robustness. In this survey,  we present a comprehensive overview of methods that improve the robustness of computer vision models against common corruptions. We categorize methods into three groups based on the model components and training methods they target: data augmentation, learning strategies, and network components. We release a unified benchmark framework (available at \url{https://github.com/nis-research/CorruptionBenchCV}) to compare robustness performance across several datasets, and we address the inconsistencies of evaluation practices in the literature. 
Our experimental analysis highlights the base corruption robustness of popular vision backbones, revealing that corruption robustness does not necessarily scale with model size and data size. Large models gain negligible robustness improvements, considering the increased computational requirements. To achieve generalizable and robust computer vision models, we foresee the need of developing new learning strategies that efficiently exploit limited data and mitigate unreliable learning behaviors. 

\end{abstract}






\IEEEpeerreviewmaketitle

\section{Introduction}

	\label{sec:introduction}
 \subsection{Background and motivation}
 Over the past decade, deep neural networks (DNNs) have been widely applied  in many fields, such as face verification (e.g. banking~\citep{Vishnuvardhan2021}, criminal tracking~\citep{9596066}, border control~\citep{8489113}), autonomous driving~\citep{7410669}, medical imaging~\citep{zhang_wu_sheng_2016} and computer-aided diagnoses~\citep{bhatt_2020}, recommender systems~\citep{3285029}, camera-based product quality monitoring~\citep{4418522}. 
 In computer vision, DNNs are a core component of hallmark tasks such as object recognition~\citep{8320684} and detection~\citep{Liu2020}, image segmentation~\citep{200105566}, captioning~\citep{210706912}, super-resolution~\citep{190206068},  reconstruction~\citep{Ben_Yedder_2020}, generation~\citep{math10152733}, and retrieval~\citep{9432822}. 
 
 However, DNNs experience performance degradation in case of data distribution shifts, where the characteristics of test data differ from those of training data~\citep{210813624}. These distribution shifts are categorized into three types: covariate shift~\citep{MaiaPolo2023}, label shift~\citep{pmlr-v202-garg23a}, and concept shift~\citep{Rostami_Galstyan_2023}. They correspond to changes in the distribution of input images and features, the distribution of labels, and the mapping relationship between images and labels, respectively. This survey focuses on out-of-distribution (OOD) generalization in covariate shifts settings, specifically those caused by common corruptions that change the appearance of input images. 
 
 A typical example of covariate shift occurs in medical image analysis, where a DNN-based  model trained on images from one hospital may not perform accurately when applied to images from another hospital (using different equipment)~\citep{Yu2018}. Unlike  doctors, who can make diagnoses using images from different hospitals, DNNs often fail to generalized in such scenarios. This issue becomes evident when highly-performing models on benchmarks are applied to real-world tasks, raising concerns about their reliability~\citep{Geirhos2018}.  The human visual system, in contrast, robustly handles unexpected or unknown cases~\citep{pmlr-v97-recht19a}.  
 
Robustness to covariate shifts is a commonly discussed topic, with most studies focusing on domain adaptation~\citep{10075484} and domain generalization~\citep{Zhou_2022}. In domain adaptation, models are tuned to a target domain using known information about it, while in domain generalization, the target domain distribution is unknown. 
Corruption robustness can be considered as a sub-problem of domain generalization~\citep{Zhou_2022}, specifically single-source domain generalization, where models are trained on one domain and tested on other unknown domains~\citep{Zhou_2022,hendrycks2019robustness}. However, it has different evaluation protocol from that of domain generalization. The datasets benchmarking corruption robustness are built on top of the original evaluation dataset with controlled corruption effects, strictly measuring the impact of corruptions. In common domain generalization, models are usually evaluated on datasets with different domain information, e.g styles, angles, etc., which do not have strict one-to-one mapping relationship to the original evaluation dataset.  Corruption robustness is evaluated by specifically designed metrics measuring the impact of corruption strength on model robustness~\citep{hendrycks2019robustness}, which is not an aspect considered in domain generalization. Moreover, evaluating corruption robustness does not involve selecting model checkpoints, architecture and hyperparameters based on their validation performance on an image subset with identical distribution to those of the unseen domains. 

Hereby we introduce the problem definition of corruption robustness from \citep{hendrycks2019robustness} and describe different common corruptions and their causes. 

\textbf{Common corruptions}
 are visual distortions caused by non-ideal imaging environments, resulting in appearance changes and data distribution shifts from the original training data. They differ from adversarial noise, which induces imperceptible changes in input images.
In~\citep{hendrycks2019robustness}, common corruptions are categorized into four groups: noise, blur, digital transformations, and weather conditions. For instance, CMOS image sensors are sensitive to different noise sources, such as photon shot noise and amplifier noise, especially under low-light conditions~\citep{BIGAS2006433}. Real-world image capture can result in motion blur due to object movement or defocus blur if the object is not in the focal plane. Image editing and compression (e.g. JPEG compression) can also affect model performance. Moreover,  extreme weather conditions,  such as rain and snow,  can significantly degrade image quality, and consequently model performance.

\textbf{Definition of  corruption robustness.}
Given a set of corruption functions $\mathcal{C}$, a classifier ${f:\mathcal{X}\rightarrow \mathcal{Y}}$  is trained on samples from distribution $\mathcal{D}$ which does not contain any corruptions in $\mathcal{C}$. Corruption robustness of models is judged by the average performance of models when classifying low-quality inputs with all the corruption effects in $\mathcal{C}$~\citep{hendrycks2019robustness}, i.e. $\mathds{E}_{c\sim \mathcal{C}}[P_{(x,y)\sim \mathcal{D}}(f(c(x))=y)]$.



 




\begin{figure}
    \centering
    \includegraphics[width=0.8\linewidth]{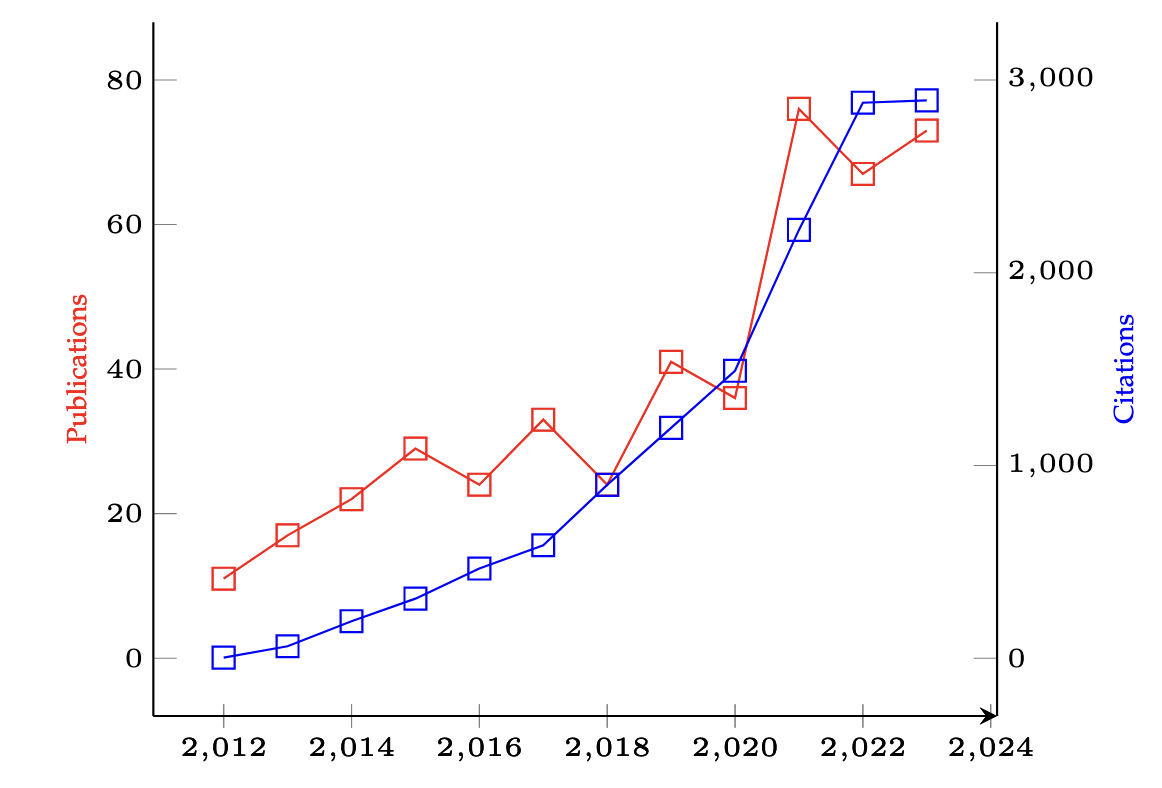}
    \caption{The number of related publications to robustness to common image corruptions and the corresponding citation numbers from the year 2012 to 2023 (searched with keywords: `image corruption', `corruption robustness', `robustness to corruption', excluding `label noise' and `noisy labels', and generated from \href{https://www.webofscience.com/wos/woscc/summary/dd9274d2-a790-4ef0-85e7-0405b1f4152c-4ed603b7/date-descending/1}{Web of Science}).}
    \label{fig:publication_years}
\end{figure}

\subsection{Comparison with other surveys }
This paper focuses  on the robustness of model predictions under image corruptions, differently from survey papers that either  benchmark general OOD performance~\citep{liu2023comprehensive} or review methods addressing domain generalization~\citep{Zhou_2022}. While~\citep{liu2023comprehensive} primarily compares different training paradigms and architectures, we provide a comprehensive taxonomy of methods addressing corruption robustness and conduct extensive evaluations of several vision backbones, from earlier CNNs to recent transformer-based foundation models, on corruption benchmark datasets beyond ImageNet-$\mathrm{C}$. We furthermore analyze how the size of pre-training datasets affects robustness of models.   
Compared to the survey that focuses on different domain generalization problem~\citep{Zhou_2022}, our review specifically targets methods addressing corruption robustness, a sub-problem of domain generalization, evaluated under the protocol of~\citep{hendrycks2019robustness}.

    \subsection{Contributions}
    We  survey research works on improving the robustness of computer vision models to common corruptions encountered in real-world scenarios, focusing on the years 2012-2024 (see~\cref{fig:publication_years} for the increasing research interests). 
    Popular pre-trained vision backbones fine-tuned for downstream tasks such as classification demonstrate outstanding accuracy performance. However, their corruption robustness is often neglected, which should be a relevant criterion for vision backbone selection since it affects model performance on downstream tasks. 
    Thus, we benchmark the robustness of popular pre-trained backbones to common corruptions and provide a comprehensive analysis and unified comparison of their performance on publicly available dataset.
    
   \noindent Our contributions are as follows: 
    \begin{enumerate}
        \item We present a  systematic overview and  a  taxonomy of works that characterize or improve the robustness of computer vision models against common corruptions. 
        We categorize methods for improving corruption robustness into three groups, namely: 1) data augmentation, 2) learning strategies, and 3)  network components.
        \item We benchmark relevant backbones for computer vision, ranging from earlier CNNs to recent foundation models, on the ImageNet-$\mathrm{C}$, ImageNet-$\mathrm{\bar{C}}$, ImageNet-$\mathrm{3DCC}$, and ImageNet-$\mathrm{P}$ datasets. We study the impact of model parameters and data size on corruption robustness, showing that scaling up models and data is not efficient for improving robustness considering the associated computational costs. The efforts made in this work for the unification of benchmark results provide a solid ground for vision backbone selection, considering both accuracy and robustness.
        \item We identify future research challenges and promising research directions for improving corruption robustness, contributing to the development of more robust and reliable computer vision systems that can operate effectively in real-world scenarios. 
    \end{enumerate}

    The remainder of this paper is organized as follows. In Section \ref{sec:relatedwork}, we describe other topics related to the robustness of vision backbones against distribution shifts. In Section \ref{sec:Definitions}, we introduce benchmark datasets and evaluation metrics. In Section \ref{sec:taxonomies}, we provide a taxonomy of methods designed to address corruption robustness and give an overview of the results of these methods (collected from the papers). We investigate the corruption robustness of different popular computer vision backbones (with a corruption robustness evaluation framework that we made publicly available) in~\cref{sec:benchmark_results}. In~\cref{sec:future}, we discuss future challenges and opportunities for improving corruption robustness, followed by conclusions in~\cref{sec:conclusion}.

\section{Related topics}
    \label{sec:relatedwork}
    In this section, we discuss other topics related to OOD generalization, specifically domain generalization, adversarial robustness, and label noise robustness.
    
    \textit{Domain generalization.}
    It addresses the OOD generalization problem by training models with data from single or multiple source domains with distinct distributions. According to~\citep{Zhou_2022}, common approaches includes domain alignment~\citep{pmlr-v28-muandet13,Li_2018_ECCV,9412797,Ghifary_2015_ICCV,101007978,Wang_2020_CVPRCrossDomain}, data augmentation~\citep{9010347,8995481,9157002,xu2021robust},  meta-learning~\citep{pmlr-v70-finn17a,10100586072_12,du2021metanorm}, ensemble learning~\citep{Niu_2015_ICCV,1010079786_5}.
    Domain alignment learns image representations invariant to different domain characteristics. For training, it requires data from multiple domains  or data augmentation that add specific domain information. Meta learning, known as learning-to-learn, adjusts model training with an meta-optimization objective which ensures loss minimization in both training and (virtal) test domain. 
    Ensemble learning is famous for learning the same model using different configurations, and combining the predictions of the ensemble models was shown to improve generalization performance. 
    Corruption robustness is a problem within the field of domain generalization: we focus on methods beyond the taxonomy provided in~\citep{Zhou_2022}, which are evaluated under the specific protocol proposed by~\citep{hendrycks2019robustness}.

 \textit{Adversarial robustness.}
    DNNs are  vulnerable to small, imperceptible perturbations of their inputs, leading to misclassifications~\citep{181000069}. There are naturally adversarial samples~\citep{hendrycks2021nae} and artificial samples generated by adversarial attacks~\citep{goodfellow2015explaining}. Effective adversarial noise depends on the knowledge of a specific model, e.g. white-box attacks with access to the model or black-box attacks without knowledge of the model~\citep{181000069}. This is different from common corruptions that can naturally occur in any environment and do not depend on specific model architectures. Methods to improve adversarial robustness include adversarial training~\citep{Balunovic2020Adversarial,BS_2020_CVPR},  adversarial example detection~\citep{Qin2020Detecting,Deng_2021_CVPR}, certified defence~\citep{jia2020certified,NEURIPS2020_5fb37d5b}, and input transformations~\citep{Gupta_2019_ICCV,Yuan_2020_CVPR}. In~\citep{kireev2021effectiveness}, adversarial noise for training has been shown to benefit the robustness of models to common image corruptions. 

    \textit{Label noise robustness.} 
    Label noise refers to incorrect ground truth labels that  mislead the training of models, resulting in sub-optimal model performance. Label noise is common due to non-expert labeling~\citep{Zhang2016} and ambiguous data that can make accurate annotation challenging, 
    or pseudo labels created by pre-trained models~\citep{9086055}.  Detecting and mitigating label noise are essential to improve classification performance and robustness of models.  Related methods include ground truth label estimation~\citep{NIPS2012_cd00692c,6823124}, direct learning with label noise~\citep{6878424}, sample selection~\citep{NEURIPS2018_a19744e2,Wei_2020_CVPR}, sample weighting~\citep{Wang_2018_CVPR,pmlrv119harutyunyan20a},  meta-learning~\citep{NEURIPS2019_e58cc5ca} and etc.

\begin{table*}
\tiny
	\centering                       
	\caption{Benchmark datasets for corruption robustness evaluation, containing visual distortions such as noise, blur, weather effects and digital transformations.}      
	\label{tab:benchmark}       
	\begin{tblr}{
            hlines,
            rowsep=1pt,
            rows = { font=\scriptsize},
            column{1} = {l}{0.26\linewidth},
            column{2} = {l}{0.65\linewidth},
            hline{1,2,10} = {-}{1pt}
            }   
		   \bfseries Dataset &  \textbf{Image variation types} \\
      CIFAR-$\mathrm{C}$~\citep{hendrycks2019robustness}
     &   \textbf{Noise} (Gaussian, Impulse, Shot, Speckle); \textbf{Blur} (Defocus, Glass, Gaussian, Motion, Zoom);
      \textbf{Weather} (Fog, Frost, Snow, Spatter); \textbf{Digital} (Brightness, Contrast, Elastic, JPEG, Pixelate, Saturate)\\
      
     ImageNet-$\mathrm{C}$~\citep{hendrycks2019robustness}
    &  \textbf{Noise} (Gaussian, Shot, Impulse);  \textbf{Blur} (Defocus, Frosted glass, Motion, Zoom); \textbf{Weather} (Snow, Frost, Fog);  \textbf{Digital} (Brightness, Contrast, Elastic, Pixelate, JPEG)
    \\
    
    CIFAR-$\mathrm{\bar{C}}$~\citep{mintun2021interaction} &  \textbf{Noise} (Brown, Blue);   \textbf{Blur} (Circular Motion Blur, Transverse Chromatic Aberration);  \textbf{Digital}  (Checkerboard, Lines, Inverse Sparkle, Sparkles, Pinch and Twirl, Ripple)
    \\
    
    ImageNet-$\mathrm{\bar{C}}$~\citep{mintun2021interaction} &  \textbf{Noise} (Brown, Perlin, Blue, Plasma, Single Frequency, Concentric Sin Waves);  \textbf{Digital} (Checkerboard,  Caustic Refraction, Inverse Sparkle, Sparkles)
    \\
      CIFAR-$\mathrm{P}$~\citep{hendrycks2019robustness}&  \textbf{Noise} (Gaussian, Shot); \textbf{Blur} (Motion, Zoom); \textbf{Weather} (Snow); \textbf{Digital} (Brightness, Rotate, Scale, Tilt, Translate)\\ 
     
      ImageNet-$\mathrm{P}$~\citep{hendrycks2019robustness}&   \textbf{Noise} (Gaussian, Impulse, Shot, Speckle); \textbf{Blur} (Defocus, Glass, Gaussian, Motion, Zoom); \textbf{Weather} (Fog, Frost, Snow, Spatter);  \textbf{Digital} (Brightness, Contrast, Elastic, JPEG, Pixelate, Saturate)\\ 
     
      ImageNet-$\mathrm{3DCC}$~\citep{kar2022d}&   \textbf{Noise} (ISO, Low-light);  \textbf{Blur} (Far focus, Near focus, XY-Motion, Z-Motion); \textbf{Weather} (Fog 3D); \textbf{Digital}  (Bit Error, Color Quantization, Flash, CRF Compress, ABR Compress)  
     \\
  
     CCC~\citep{press2022ccc}&  \textit{Appied in random pairs:}  \textbf{Noise} (Gaussian, Impulse, Shot, Speckle);   \textbf{Blur} (Defocus, Glass, Gaussian, Motion, Zoom); \textbf{Weather} (Fog, Frost, Snow, Spatter); \textbf{Digital} (Brightness, Contrast, Elastic, JPEG, Pixelate, Saturate) 

	\end{tblr}
\end{table*}

\section{Datasets and metrics}
	\label{sec:Definitions}

    In the context of OOD generalization, several datasets have been released to evaluate model performance. 
    OOD data can take various forms. For instance,  imperceptible adversarial noise  added to images results in distribution shifts that impair classification performance. In~\citep{hendrycks2021nae}, two benchmark datasets named ImageNet-$\mathrm{A}$ and ImageNet-$\mathrm{O}$ were proposed. Both datasets contain naturally occurring adversarial images which are hard to recognize. 
    
    ImageNet-$\mathrm{O}$ contains images of unknown classes while ImageNet-$\mathrm{A}$ contains images of known classes, i.e. assumed to have been seen during training on ImageNet-1K. As found in~\citep{li2021rethinking} that the large, unusual background is an essential reason for misclassification, ImageNet-A-plus~\citep{li2021rethinking} with reduced background influence  was proposed.  In~\citep{220304412}, the authors proposed ImageNet-Patch that  contains images with adversarial patches that lead to degraded performance.   
    
	Popular datasets to test OOD generalization are OOD-CV~\citep{211114341}, SI-Score~\citep{yung2022siscore}, WILDS~\citep{201207421}, and ImageNet-$\mathrm{E}$~\citep{Li_2023_CVPR}. They contain variations like pose, shape, texture, context, location, size, rotation, etc. Images with different renditions from the training data are considered OOD data as well. Typical datasets are ImageNet-$\mathrm{R}$~\citep{hendrycks2021faces}, ImageNet-$\mathrm{D}$~\citep{210412928}, ImageNet-Cartoon~\citep{salvador2022imagenetcartoon}, ImageNet-Sketch~\citep{wang2019learning}, and ImageNet-Drawing~\citep{salvador2022imagenetcartoon}. These datasets allow for evaluating the generalization ability of models to different renderings and the dependence of models on natural textures.

    Within the scope of this survey, we review  benchmark datasets specifically designed to evaluate the corruption robustness of classification models (see~\cref{tab:benchmark} for an overview), and  corresponding evaluation metrics.

      \begin{figure*}[!t]
    \centering
     \subfloat[]{
     \includegraphics[width=0.2\linewidth]{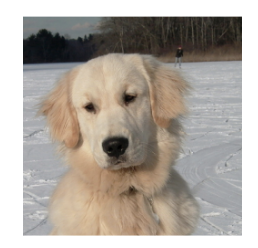}
     \label{fig:orgdog}}
    \subfloat[]{
    \includegraphics[width=0.8\linewidth]{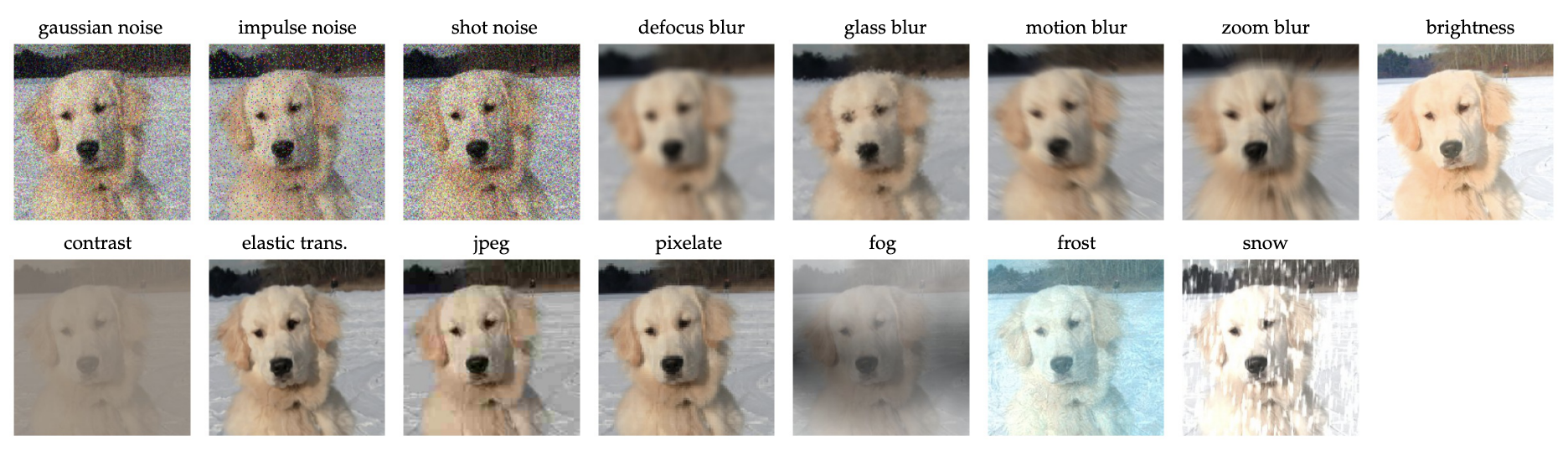}
    \label{fig:cifar10c}}
    
    \subfloat[]{
    \includegraphics[width=0.5\linewidth]{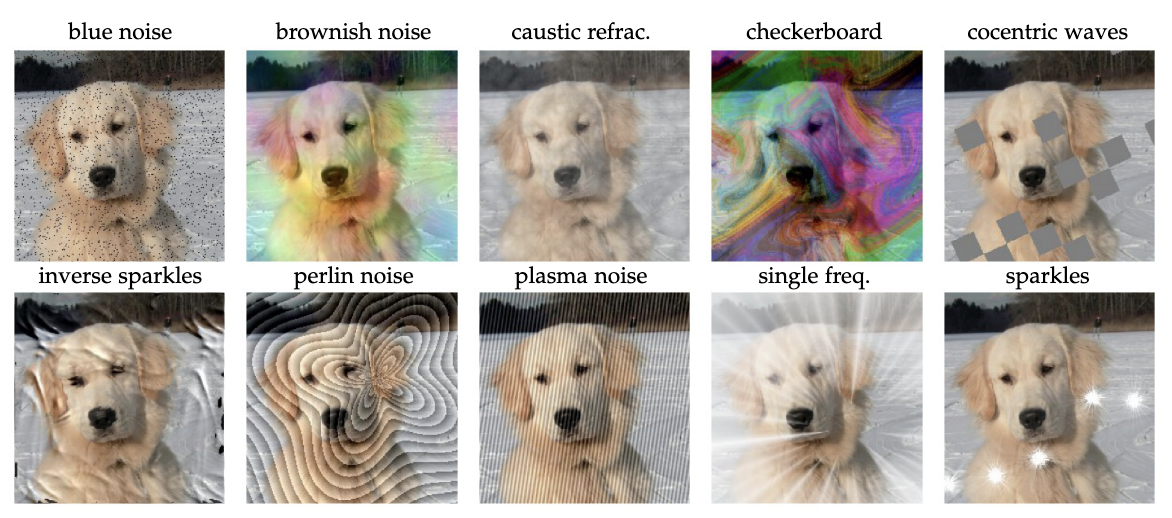}
    \label{fig:cifar10cbar}} \hspace{2em}
    \subfloat[]{
    \includegraphics[width=0.3\linewidth]{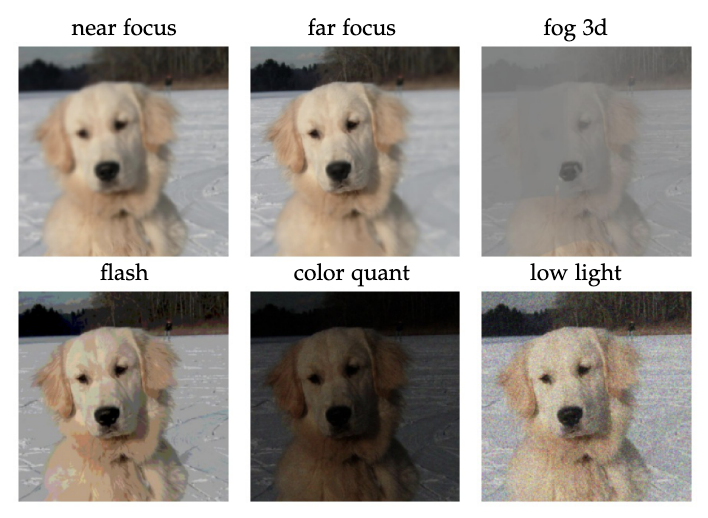}
    \label{fig:imagenet3dcc}}

    \caption{(a) An image from ImageNet and its corrupted versions from (b) ImageNet-$\mathrm{C}$, (c) ImageNet-$\mathrm{\bar{C}}$ (perceptually different from ImageNet-$\mathrm{C}$), and (d) ImageNet-$\mathrm{3DCC}$ (using 3D geometric information to improve realness).}
    
\end{figure*}

    \subsection{Benchmark datasets for corruption robustness}
    \label{sec:benchmark}
    Two benchmark datasets,  CIFAR-$\mathrm{C}$ and ImageNet-$\mathrm{C}$, were proposed in~\citep{hendrycks2019robustness}. They were generated from the validation set of CIFAR and ImageNet, respectively. CIFAR-$\mathrm{C}$ contains 19 corruptions while ImageNet-$\mathrm{C}$ contains 15 corruptions. Four extra corruptions for ImageNet-$\mathrm{C}$ serve as a simplified robustness validation set. Each corruption has five levels of severity. High severity indicates high strength of corruption applied to the original images. \cref{fig:cifar10c} shows the corruption types proposed in~\citep{hendrycks2019robustness} (at severity 3).

    The corruptions in CIFAR-$\mathrm{C}$ and ImageNet-$\mathrm{C}$ do not completely cover variations occurring in the real world. Two complementary datasets, CIFAR-$\mathrm{\bar{C}}$ and ImageNet-$\mathrm{\bar{C}}$ were thus designed~\citep{mintun2021interaction}, that contain the 10 most perceptually dissimilar corruptions to those in CIFAR-$\mathrm{C}$ and ImageNet-$\mathrm{C}$ (see~\cref{fig:cifar10cbar}).   As shown in~\citep{taori2020measuring}, previous synthetic datasets fail to assess real OOD generalization performance. 
    To ensure realness of images that possibly occur in real cases, ImageNet-$\mathrm{3DCC}$ includes 3D corruptions that consider the scene geometry (see~\cref{fig:imagenet3dcc})~\citep{kar2022d}. For more data variety, the test images in Continuously Changing Corruptions (CCC) dataset~\citep{press2022ccc} are generated by combining pairs of corruptions.  Corruption types of varying severity and characteristics are combined through random walks to create a diverse set of benchmarks. 
    To inspect how model predictions change as the effect of corruptions varies over time, time-dependent corruption sequences were generated in the CIFAR-$\mathrm{P}$ and ImageNet-$\mathrm{P}$ datasets~\citep{hendrycks2019robustness}. Each frame of a sequence is corrupted based on its previous frame.

    Synthetic benchmarks offer extensive ground for comparing the robustness of computer vision models. These datasets contain a wide range of corruption types, enabling the assessment of different aspects of robustness and generalization that are not typically evaluated when testing new algorithms and models. We thus contend that incorporating robustness testing into  performance evaluation of computer vision models should be a standard practice.

    \subsection{Evaluation metrics}
    \subsubsection{Mean corruption and relative corruption error }
    Several metrics to measure corruption robustness were proposed in~\citep{hendrycks2019robustness}. The mean corruption error ($\mathrm{mCE}$) and relative corruption error ($\mathrm{rCE}$) are basic metrics, computed as: 
    \begin{equation}
    \label{equ:1}
    \mathrm{mCE} = \frac{1}{|C|}\sum_{c \in C }\frac{\sum_{s=1}^5 E_{s,c}^f}{\sum_{s=1}^5 E_{s,c}^{baseline}},
    \end{equation}
\noindent and 
    \begin{equation}
    \label{equ:rmce}
    \mathrm{rCE} = \frac{1}{|C|}\sum_{c\in C}\frac{\sum_{s=1}^5 (E_{s,c}^f-E_{original}^f)}{\sum_{s=1}^5 (E_{s,c}^{baseline}-E_{original}^{baseline})},
    \end{equation}
    where $C$ is the set of corruptions in the test set and $E_{s,c}^f$ is the classification error rate of a tested model $f$ for a type of corruption $c$ with severity $s$ ranging from one to five. Before aggregating a classifier's performance across different corruptions, both metrics normalize the classification error of each corruption by the error of a ${baseline}$ model. This comprehensively evaluates corruption robustness, as different corruptions induce different difficulties in classification.    The $\mathrm{mCE}$ measures the classification performance of a model with respect to that of the baseline over all types of image corruption in the considered test set. When $\mathrm{mCE}$ is larger than one, the model $f$ is less robust than the baseline, and vice versa. 
    The $\mathrm{rCE}$ measures the performance degradation of model $f$ on corrupted images w.r.t. their original images, and compares it with that of the baseline. Thus, a lower $\mathrm{rCE}$ value indicates greater corruption robustness compared to the baseline.

    \subsubsection{Mean flip rate}
    Mean flip rate ($\mathrm{mFR}$)~\citep{hendrycks2019robustness} evaluates the effects that consecutive corruptions have on the classification performance of models over time. It computes the probability that models change their prediction in consecutive frames~\citep{hendrycks2019robustness}. It is specifically used for image sequences in ImageNet-$\mathrm{P}$, where frames are corrupted by applying the same type and level of corruption on top of the previous frame.
    The Flip  Probability ($\mathrm{FP}$) of the tested model $f$ for the corruption $c$ is computed as: 
    \begin{equation}
        \mathrm{FP}^f_c = \frac{1}{m(n-1)}\sum_{i=1}^m \sum_{j=2}^n \mathds{1}(f(x_j^{(i)})\neq f(x_{j-1}^{(i)})),
    \end{equation}
    where $m$ is the number of corruption sequences, $n$ is the number of frames in a sequence, $x^{(i)}$ indicates the $i^{th}$ sequence in the dataset and $x_j$ indicates the $j^{th}$ frame of a sequence. $\sum_{j=2}^n \mathds{1}(f(x_j^{(i)})\neq f(x_{j-1}^{(i)}))$ measures the number of frames in a sequence having different predictions from  their previous frame, i.e. it compares the prediction of the model $f$ on the frame $x_j$ of the $i^{th}$ sequence with that on the previous perturbed images in the sequence. If the predictions are the same, $\mathds{1}(f(x_j^{(i)})\neq f(x_{j-1}^{(i)}))$ equals to zero, and the performance of the model is not affected by the considered corruptions.
     For a sequence corrupted by noise, the predictions are compared with those of the first frame, as noise is not temporally related. 
    The $\mathrm{mFR}$ is obtained by averaging the flip rate $\mathrm{FR}_c^f$ across all the corruptions:
    \begin{equation}
        \mathrm{mFR} = \frac{1}{|C|}\sum_{c\in C}\mathrm{FR}_c^f  = \frac{1}{|C|}\sum_{c\in C} \frac{\mathrm{FP}^f_c}{\mathrm{FP}^{baseline}_c},
    \end{equation}
    where $C$ is the set of corruptions and $\mathrm{FR}_c^f$ is the standardized flip probability of the model $f$ compared to that of the baseline model. The value of $\mathrm{mFR}$ is expected to be close to zero for a robust model.
    
    \subsubsection{Mean top-5 distance}
    Mean top-5 distance (mT5D) also measures robustness to time-dependent corruptions in image sequences~\citep{hendrycks2019robustness}. If a model is robust to the added corruptions, then the top-5 predictions of frames over a sequence should not be shuffled. If the top-5 predictions are shuffled w.r.t. those of the previous frames in a sequence, the model is penalized for its lack of robustness in maintaining accurate predictions under consecutive corruptions.     
    The top-5 distance over each frame of $m$ sequences belonging to a corruption $c$ is:
    \begin{equation}
        \mathrm{T5D}_c^{f} = \frac{1}{m(n-1)}\sum_{i=1}^m\sum_{j=2}^n d(\tau(x_j),\tau(x_{j-1})), 
    \end{equation}
    where $d(\tau(x_j),\tau(x_{j-1}))$ measure the deviation between the top-5 predictions of two consecutive frames, using:
    \begin{equation}
        d(\tau(x_j),\tau(x_{j-1})) = \sum_{i=1}^5 \sum_{j=min\{i,\rho(i)\}+1}^{max\{i,\rho(i)\}}\mathds{1}(1 \leq j-1 \leq 5),
    \end{equation}
    with 
    \begin{equation}
        \rho(\tau(x_{j})(k)) =  \tau(x_{j-1})(k),
    \end{equation}
    where $\tau(x_j)$ is the ranking of predictions for a corrupted frame $x_j$ and $\tau(x_j)(k)$ indicates the rank of the prediction being $k$. If $\tau(x_j)$ and $\tau(x_{j-1})$ are identical, then $d(\tau(x_j),\tau(x_{j-1})) = 0$.
    Averaging the $\mathrm{T5D}$  normalized by that of the baseline over all corruptions corresponds to 
    \begin{equation}
        \mathrm{mT5D} =  \frac{1}{|C|}\sum_{c\in C} \frac{\mathrm{T5D}_c^{f}}{\mathrm{T5D}_c^{baseline}}.
    \end{equation}

\subsubsection{Expected calibration error}
    Instead of comparing models with a baseline, expected calibration error (ECE)~\citep{04599} measures the accuracy and the confidence of the predictions, examining whether the models give reliable predictions. It is computed as: 
    \begin{equation}
        ECE = \sum_{m=1}^M \frac{|B_m|}{n}|acc(B_m)-conf(B_m)|,
        \label{equ:ece}
    \end{equation}
    where $M$ is the number of interval bins separating the predictions based on their confidence value, $B_m$ is the samples in a confidence interval bin, $n$ is the total number of samples in the test set, $acc(B_m)$ is the test accuracy on the set $B_m$ while $conf(B_m)$ is the average confidence of the corresponding predictions. The lower the ECE, the more reliable the predictions. Instead of being a result-oriented metrics, ECE provides indications on how reliable the predictions are when facing OOD data.

    \section{Methods for improving corruption robustness}
    \label{sec:taxonomies}
    \label{subsec:tax_corrupt}
\begin{figure}[!t]
	    \centering
	    \includegraphics[width=0.8\linewidth]{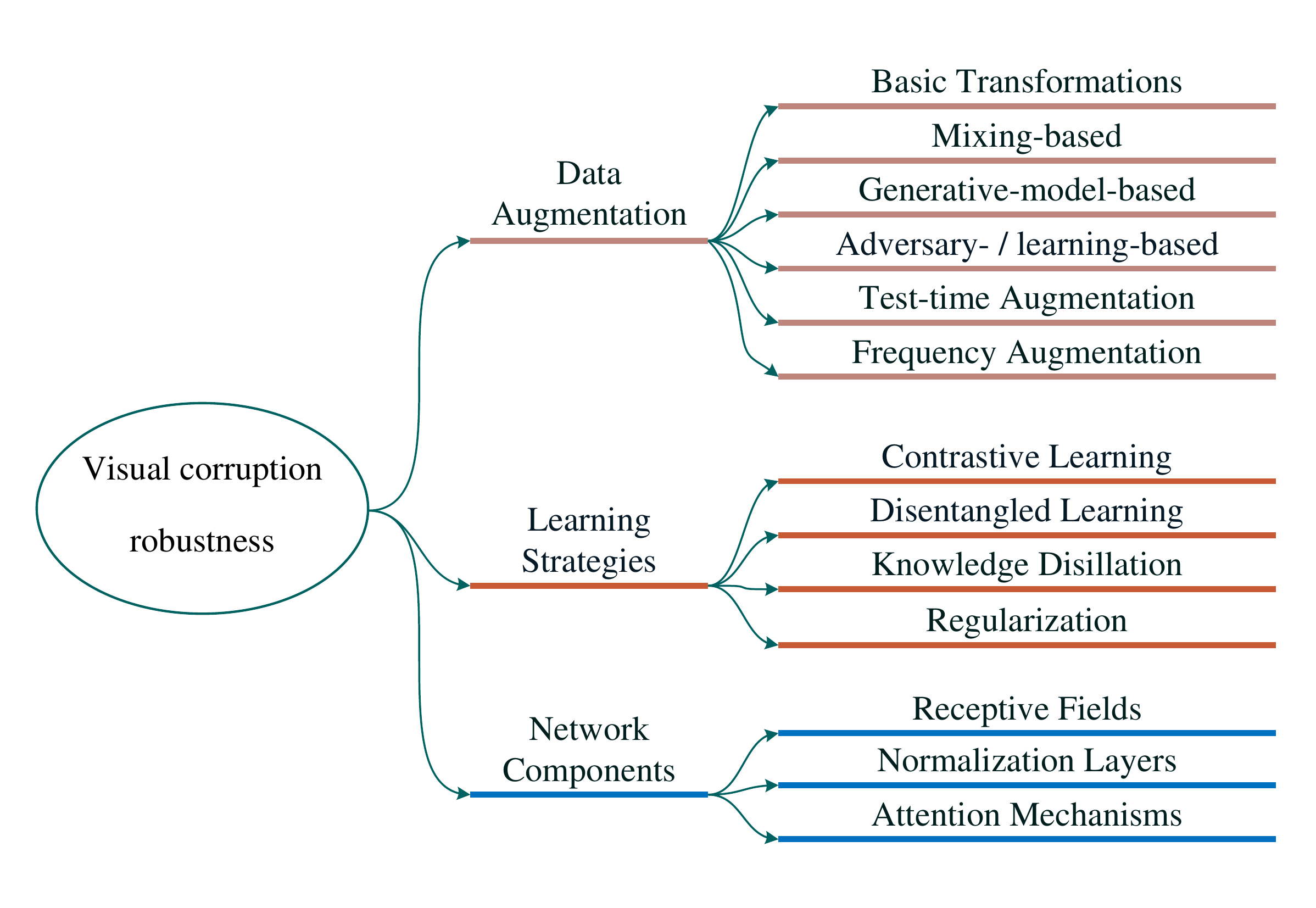}
	    \caption{Taxonomy of methods improving corruption robustness.}
	    \label{fig:overall_structure}
	\end{figure}

    We present a systematic overview of  methods to improve corruption robustness, categorized into three groups based on the addressed part of the models and training methods. These categories are  1) data augmentation, 2) learning strategies, and 3) network components.  The taxonomy is shown in~\cref{fig:overall_structure} and an overview of methods is provided in~\cref{tab:taxonomy}.  We collect the benchmark results of these methods from their original papers.

     \begin{table*}
    \tiny
	\centering                      
	\caption{Overview of methods addressing corruption robustness, organized in three categories: data augmentation, learning strategies, and network components. }         
	     \label{tab:taxonomy} 
     \SetTblrInner{rowsep=1pt}
	\begin{tblr}{
        hlines,
        rows = { font=\scriptsize},
        cell{2}{1} = {r=34}{},
        cell{36}{1} = {r=13}{},
        cell{49}{1} = {r=12}{},
        cell{2}{2} = {r=8}{},
        cell{10}{2} = {r=10}{},
        cell{21}{2} = {r=9}{},
        cell{30}{2} = {r=6}{},
        cell{36}{2} = {r=6}{}, 
        cell{42}{2} = {r=2}{}, 
        cell{45}{2} = {r=4}{},
        cell{49}{2} = {r=6}{},
        cell{55}{2} = {r=2}{},
        cell{57}{2} = {r=4}{},
        column{1} = {l}{0.09\linewidth},
        column{2} = {l}{0.08\linewidth},
        column{3} = {l}{0.14\linewidth},
        column{5} = {l}{0.52\linewidth},
        column{4} = {c},
        hline{1,2,61} = {-}{1pt}
        }     
		\bfseries Category &\bfseries Sub-cat. &\bfseries Method&\bfseries  Codes &\bfseries Remark  \\ 
		\textbf{Data \newline augmentation} 
    & Mixing  & AugMix~\citep{hendrycks2020augmix} & Yes  & Combine augmentations randomly with JS-Divergence, ensuring consistency of images  \\  
    &  & Targeted Aug. ~\citep{gao2022outofdistribution}&  No  &  Copy-paste the predictive features of images that may vary across different domains  \\ 
     & &  LISA~\citep{pmlr-v162-yao22b}&   Yes &  intra-label or intra-domain augmentation \\  
     & & Mixup~\citep{mixup}&  Yes  &   Linearly interpolate input examples and their ground truth \\   
    &    &  PixMix~\citep{Hendrycks_2022_CVPR}&  Yes & Use natural structural complexity of e.g. fractals pictures \\
      &    & SmoothMix~\citep{Lee_2020_CVPR_Workshops}  & Yes  &  Reducing strong edge when two images are mixed together\\
     &    &  GuidedMixup~\citep{Kang_Kim_2023} & Yes  & Use saliency map to guide mixup strategy \\
      &    & PuzzleMix~\citep{pmlr-v119-kim20b}  & Yes  & Mix image patches that are the most salient \\
        
    & Adversary-/ learning-based  & AutoAug.~\citep{cubuk2019autoaugment} & Yes & Apply reinforcement learning to learn  the best augmentations  \\ 
    & & Adv. AutoAug.~\citep{zhang2020adversarial} & No & Adversarial learning augmentation policies with reinforcement learning \\
    &  & PRIME~\citep{prime}& Yes   &  Augmentations with maximum entropy are selected during training  \\  
 &  & ANP~\citep{Liu2021}&  Yes  &  Adversarial noise is injected into hidden layers   \\  
  &Adversarial  &AugMax~\citep{augmax}& Yes   &  Adversarially select augmentations and their combined weights \\   
  &  & ME-AdA~\citep{NEURIPS2020_a5bfc9e0}& Yes   & Adversarial training with  a maximum-entropy-based tern to regularize the distribution of data \\  
   &  & RSPC~\citep{Guo_2023_CVPR} & Not yet  &  Reduce sensitivity to patch corruptions   \\
  &  &  GAT~\citep{9301856}&  No  & Gaussian noise generate adversarial examples       \\ 
  &  & NoisyMix~\citep{NoisyMix}&  No  &  Apply adversarial noise in both input and feature space  \\  
  &   &DAJAT~\citep{effectiveaug}&  Yes  &  Common augmentations combined with adversarial training   \\  

 & Generative  & MDA~\citep{robey2020modelbasedrobustdeeplearning}&  Yes  &  Model-based augmentation with natural effect\\ 

 & Test-time \newline augmentation & Test-time Aug.~\citep{NEURIPS2020_2ba59664}&  Yes  &  Augmentations to use are selected based on the characteristics of test data  \\  
  &    &  DDA~\citep{Gao_2023_CVPR} &  Yes  & Align target data to source domain \\
   &    &  selective-TTA~\citep{SON2023119148} & No & Apply augmentation when the uncertainty of predictions of inputs is high \\
   &    &  GPS~\citep{molchanov2020greedy} & Yes  & Learn test-time augmentation policies  \\
    &    &   TTA aggregation~\citep{shanmugam2021better} &  Yes &  Learn to combine augmentations \\
   &    &  Adaptive-TTA~\citep{AdaptiveTTA} & No  &  Apply temperature scaling to adjust the weights of predictions \\
   &    & TeSLA~\citep{Tomar_2023_CVPR} &  Yes & Self-learning adversarial augmentation  \\
   &    & Cyclic-TTA~\citep{pmlr-v180-chun22a} &  No & Cyclic policy search for augmentation operations  \\
    &    & SuccessTTA~\citep{10448390} & No  &  Successively predict transformations by one forward pass  \\

 &  Frequency  & APR-SP~\citep{Chen_2021_ICCV} &  Yes & Mixing amplitude information of images while preserving phase information \\
 &    & VIPAug~\citep{Lee_Lee_Myung_2024} &  Yes & Apply weak variations to vital phases and strong variations to non-vital phases \\
  &    &  AFA~\citep{vaish2024fourierbasis} & Yes  & Apply Fourier-basis functions to image spectrum \\
 &    & HybridAugment++~\citep{Yucel_2023_ICCV}  & Yes  & Mix high-frequency spectrum of images and remain the phase information \\
  &   &AugSVF~\citep{220212412}& No   &  Expands the options of operations with Fourier noise in AugMix framework  \\    &  &RobustMix~\citep{2022robustmix}& No   &  Frequency bands of input samples are mixed with energy-dependent weights  \\

     \textbf{Learning \newline strategies} &  Contrastive 
          &  SimCLR~\citep{chen2020simple}&  Yes  &  Self-supervised contrastive learning \\ 
   &  & G-SimCLR~\citep{chakraborty2020gsimclr}& Yes  & Pseudo labels to avoid representation of the same category to be distant  \\  
  &  & SupCon~\citep{khosla2021supervised}& Yes & Ground truth labels  to avoid representation of the same category to be distant \\  
 &   & Balanced-SupCon~\citep{pmlr-v162-chen22d}&  Yes  &  Apply InfoNCE loss to avoid class collapse  \\  
  &  &    Percept.-similar~\citep{210606620}&  No  &  Representations of perceptually similar images are learned to be similar   \\ 
  &    &  Adv. Con~\citep{NEURIPS2020_ba7e36c4}&  Yes  &  Contrastive learning with adversarial noise  \\ 
  
    &  Knowledge \newline distillation  &  NoisyStudent~\citep{9156610}& Yes   & A network trained on cleaned data teaches a student network trained on unlabeled data \\
  &  & Aux. training~\citep{Zhang_2020_CVPR}&  Yes  & Auxiliary classifier trained to predict corrupted data correctly, implicitly enforcing invariance    
     \\ 

& Disentangled   & Unpaired synthesis~\citep{du2020learning} & Yes  & Learn content and style codes separately \\

      & Regularization   & RoHL~\citep{Saikia_2021_ICCV} & No  & Minimize the total variation of activations of convolutional layer \\
   &    & FPCM~\citep{Bu_2023_ICCV} & Yes  & Frequency control module to learn model preference towards low- and high-frequency \\
   &    & JaFR~\citep{ijcai2022p93} & Yes  & Regularize the Jacobians of models to have a larger ratio of low-frequency components \\
   &    & SAM~\citep{foret2021sharpnessaware} &  Yes & Smoothing the sharpness of loss landscape \\
   
  \textbf{Network \newline  components}  
 & Receptive  \newline field &  Push-pull Layer~\citep{RN124}&  Yes  &  Biologically-inspired kernels result in robust response  \\  
       &    &  BlurPool~\citep{pmlr-v97-zhang19a} & Yes  & Apply low-pass filter before downsampling to avoid aliasing \\
 &  &  On-off Kernels~\citep{210607091}&  Yes  &  Gaussian kernels with opposite directions  \\  
 &   & Smoothing Kernel~\citep{pmlr-v162-park22b}&  Yes  &  Kernel results in stabilized feature maps  \\  
 &   & Winning hand~\citep{winninghand}& No   &  Compressed networks from over-parameterized models  \\  
 &  &  LCANet~\citep{pmlr-v162-teti22a}& Yes   &    Apply sparse coding layers in the front end       \\

  & Normalization \newline layer &   AdaBN~\citep{benz2021revisiting}&  No   & Adapt BN statistics to fit new dataset \\ &    & CorrectBN~\citep{NEURIPS2020_85690f81}  & Yes  &  Correct BN statistics from both training and test data \\

 & Attention   \newline mechanism  & ViT~\citep{paul2021vision}  & Yes  & Vision transformer containing attention mechanism  \\
 &   & Patchifying Input~\citep{wang2023can}   & Yes  & Replacing down-sampling block by that of vision transformers \\
 &    &  FAN~\citep{pmlr-v162-zhou22m}   &  Yes  &  Fully attentional networks considering self-attention and channel attention   \\   
 &   & RVT~\citep{Mao_2022_CVPR}  & Yes & Consist of robust building blocks e.g. patch embedding.\\

	\end{tblr}
\end{table*}

    \subsection{Data Augmentation}
    Recent evidence suggests that the robustness of models benefits from well-annotated, large-scale datasets since they contain a wide range of variations that might occur in the real world~\citep{9156610,taori2020measuring}.  However, obtaining such datasets is expensive. Thus, researchers generate synthetic data using different augmentations to increase data variety and bridge the distribution gap between training and test data~\citep{hendrycks2020augmix,robey2020modelbased}. This is shown to improve the corruption robustness of vision models~\citep{robey2020modelbased}.

     \textit{Basic transformations.} Basic data augmentations consist of simple image transformations such as flipping, cropping, rotation, translation, color jittering, edge enhancement, etc~\citep{RN116}. Kernel filters can blur images to mimic motion blur or defocus blur~\citep{kang2017patchshuffle}. Pixel erasing removes pixels randomly from images~\citep{zhong2017random}, inspired by dropout regularization~\citep{JMLR:v15:srivastava14a}. Additive Gaussian and speckle noise applied to images also improves model corruption robustness~\citep{10.1007/978-3-030-58580-8_4}. Instead of augmenting an entire image, Patch Gaussian augmentation~\citep{190602611} applies Gaussian noise to small image patches, and benefits the classification performance  on both clean and corrupted images. For vision transformers, patch-based negative augmentation, such as patch shuffling, rotation and refill, penalizes transformers that excessively rely  on local features in patches which are hard to recognize by human observers~\citep{qin2021understanding}. This encourages transformers to focus on global structural features instead. 

    \textit{Mixing-based.}  Combining different transformations further increases data variety. Mixup~\citep{mixup} generates images by linearly interpolating two source images. The ground truth of the interpolated images is also interpolated from the ground truth of the two original images. Thus, a model is trained to predict the weight of each source image in the interpolated version. This also improves robustness to label noise and adversarial attacks. Directly mixing two non-equal-size images results in strong edges, which might be exploited by models for classification during training. To address this, SmoothMix~\citep{Lee_2020_CVPR_Workshops} processes two randomly selected images by element-wise multiplying them with a mask containing a smoothing area and its complementary version, before mixing them. To avoid the mixing operation from damaging image semantics relevant to the tasks at hand, GuidedMixup~\citep{Kang_Kim_2023} and PuzzleMix~\citep{pmlr-v119-kim20b} combine the most salient parts of paired images. The former pairs images with low semantic conflict and mixes the whole images, while the latter mixes patches of two images with high saliency.
    LISA~\citep{pmlr-v162-yao22b} learns invariant predictors through selective augmentation, i.e. intra-label and intra-domain augmentations.  Intra-label augmentation interpolates samples with the same labels but in different domains, learning useful semantics rather than spurious relations between domain information and the ground truth labels. Intra-domain augmentation interpolates samples with different labels but in the same domain. The learned representations can ignore unrelated domain information and focus on discriminative information to classify samples.
    AugMix~\citep{hendrycks2020augmix} augments images by randomly selecting augmentation operations.  Jensen-Shannon divergence is applied to enforce the consistency  of the augmented images (compared with the source image). PixMix~\citep{Hendrycks_2022_CVPR} increases structural complexity of images by mixing them with fractal images. 
    Targeted augmentation~\citep{gao2022outofdistribution} uses copy-paste technique, i.e. the object to be recognized is copied and pasted to different backgrounds, to preserve predictive features that may vary across different backgrounds in the augmented images. This improves the generalization of models to OOD data, as it avoids learning the spurious correlation between backgrounds and the ground truth labels~\citep{xiao2021noise}.

    \textit{Generative-model-based.} Generative adversarial neural networks (GANs)~\citep{karras2019stylebased, RN115,mirza2014conditional, CycleGAN2017} and variational autoencoders (VAEs)~\citep{YUN2020317} are also deployed for data augmentation, e.g. mimicking corruption effects on images for training and thus improving corruption robustness~\citep{robey2020modelbasedrobustdeeplearning}. Inspired by style transfer, where different augmentation effects can be considered as different domains, CycleGAN~\citep{CycleGAN2017} utilizes a pair of autoencoders for bidirectional mapping, i.e. transferring an image from one domain to another. The cycle consistency loss ensures that images mapped back to the source domain are not contradictory to the source images, thus improving the quality of the generated images. CycleGAN was used in~\citep{9191127,8982932} to perform data augmentation on medical images. Model-based image generators aim at reducing data imbalance, rather then directly improving corruption robustness.  Although GANs and VAEs can generate an unlimited number of images with various levels of augmentation, the training process of them needs large computational resources and has high requirements of training data. Furthermore, the realness of the generated images needs more consideration, as it affects the quality of training data.
    
    \textit{Adversary- or learning-based.} Adversarial training forces models to classify images with  (imperceptible) adversarial noise correctly~\citep{9301856,madry2018towards}, inducing models to be as robust as biological neurons to adversarial perturbations~\citep{pmlr-v162-guo22d}.
    
    Adversarial training also improves the robustness of models to common image corruptions, e.g.  Gaussian adversarial training (GAT)~\citep{9301856} that enforces Gaussian distribution regularization to adversarial noise. A regularization term based on maximum entropy formulation is proposed in~\citep{NEURIPS2020_a5bfc9e0}. It encourages perturbing the source data distribution rather than the data directly. Thus, the generated adversarial samples bring more uncertainty to predictions.  
    Generating adversarial input patches for transformers improves their robustness against corruptions~\citep{Guo_2023_CVPR}.  
    
    Rather than directly adding noise to the input, adversarial noise propagation (ANP) focuses on injecting noise to the hidden layers of NNs, which can be easily combined with other adversarial training methods~\citep{Liu2021}. ANP not only improves adversarial robustness but also corruption robustness. Similarly, NoisyMix~\citep{NoisyMix} applies augmentations in both input and feature space, making the decision boundary smoother.  Diverse augmentation-based joint adversarial training  (DAJAT)~\citep{effectiveaug} further combines adversarial training with common augmentations: two models are trained with augmented images and their source images respectively.  The models share the same weights but with different statistics for batch normalization layers, becoming more robust to distribution shifts.
    
    Rather than injecting noise to images, many works focus on the adversarial combinations of different augmentation operations. The authors of~\citep{zhang2020adversarial} proposed adversarial Autoaugment. Autoaugment~\citep{cubuk2019autoaugment} is based on reinforcement learning to search for the best combination of augmentations. The adversarial version finds adversarial augmentation policies.
     However, the search process is computationally intensive and dataset-specific. AugMax~\citep{augmax} adversarially selects augmentations and calculates the combined weights  with back-propagation, expanding the AugMix framework~\citep{hendrycks2020augmix}.  PRIME~\citep{prime} selects image transformation based on whether the distributions of the chosen image transformations have maximum entropy. This guarantees that the complexity of samples reaches certain generalization bounds. However, most adversarial-based methods require extensive computations due to back-propagation process, highlighting a research gap in reducing computational costs while maintaining comparable performance.

    \textit{Test-time augmentation.}
    Test-time augmentation (TTA) applies image transformations, e.g. cropping and flipping during testing~\citep{NIPS2012_c399862d}. These methods aggregate predictions from multiple augmented version of an image, inspired by ensemble learning~\citep{10.1007/978-3-030-92185-9_46}. In~\citep{Hekler_Brinker_Buettner_2023}, the authors demonstrated that TTA benefits the calibration of the confidence value of predictions, resulting in reliable and safe decision-making models under real-world uncertainty. 
     To reduce the computational cost for aggregating predictions, Selective-TTA~\citep{SON2023119148} applies test-time augmentation  when the uncertainty of prediction of input images is high. 
     Rather than averaging the predictions of differently augmented images, the work in~\citep{shanmugam2021better} learns the weights to aggregate the predictions of differently augmented images for a final decision.  Similarly, Adaptive-TTA~\citep{AdaptiveTTA} applies temperature scaling to adjust the weights of each prediction in the final decision. 
    
     Focusing more on improving augmentation operations, augmentations selected by an auxiliary model that predicts suitable augmentations with low loss values were applied~\citep{NEURIPS2020_2ba59664}.
     TeSLA~\citep{Tomar_2023_CVPR}  uses knowledge distillation in adversarial augmentation during test-time. 
    The student (target) model is updated when  the predictions of augmented images by the student model and non-augmented versions by the teacher model are inconsistent. When the adversarial augmented images results consistent outputs as those of non-augmented images, the model is not updated and these images are considered hard samples. Through this, the student model distils information from easy samples and discards hard samples, resulting in better decision boundary separation. The authors in~\citep{molchanov2020greedy} proposed learnable test-time augmentation, through searching for augmentation policies greedily. 
    Cyclic-TTA~\citep{pmlr-v180-chun22a} improves the policy-search-based method with cyclic search. The loss predictor is trained to predict transformations that suppresses corruption effects in images. Unlike~\citep{NEURIPS2020_2ba59664}, the augmented images are fed to the loss predictor iteratively until the loss predictor performs the optimal transformation or reaches predefined maximum number of iterations, instead of the target network directly.  This allows more flexibility to the magnitude of applied transformations. Different from~\citep{pmlr-v180-chun22a} which requires multiple forward passes to predict transformations, SuccessTTA~\citep{10448390} only needs one forward pass for backbone network, with an extra RNN-based transformation predictor outputing the optimal transformation which has minimum loss consecutively.

    DDA~\citep{Gao_2023_CVPR} aligns the domain information of test images to that of training images. The authors trained an image-level visual domain prompt for target domain, which is used to generate prompted images with augmentation mapped to the source domain information. TTA, when applied during inference, further improves the corruption robustness of models that rely only on training-time methods. But it might not be suitable for scenarios requiring quick inference time, e.g. autonomous driving~\citep{7410669}.

    \textit{Frequency augmentation.}
    Most methods focus on complicating visual transformations in the spatial domain, mimicking real-world cases. However, these transformations only affect a limited range of frequency components of images, as corruption effects have fixed frequency characteristics~\citep{3454287}. Spatial augmentations might not cover all real-world corruptions. Augmenting images in the frequency domain allows to perturb a broader spectrum of frequency components, further increasing data variety. Under the AugMix framework, AugSVF~\citep{220212412} expands the options of visual transformations with additive Fourier-basis noise, allowing to perturb specific spatial frequencies.
    AFA~\citep{vaish2024fourierbasis} randomly samples Fourier-basis functions and applies them to the RGB channels of images separately, with lower computation costs compared to~\citep{220212412}.   As phase information is more robust against noises and common corruptions than amplitude information, APR-SP~\citep{Chen_2021_ICCV} mixes the amplitude spectrum of paired images, while preserves the phase information. This reduces the reliance of models on amplitude information for classification, thus improving corruption robustness. Similarly, HybridAugment++~\citep{Yucel_2023_ICCV} mixes specifically high-frequency spectrum, to reduce model reliance on high-frequency information for classification. Phase spectrum of images changes when corresponding amplitude spectrum changes.  VIPAug~\citep{Lee_Lee_Myung_2024} applies variations to phase based on whether they contain important information according to the magnitude of amplitudes (higher magnitude indicates more importance).  It adds weak variations to important phases and strong variations to non-important phases.  
    RobustMix~\citep{2022robustmix} fuses the frequency bands of two images. The weight of each frequency band depends on its relative amount of energy. Thus, the models are regularized to classify based on low-frequency spatial features, which are more robust to small changes in the inputs  than high-frequency features~\citep{wang2020high}, as they depend more on shapes rather than textures for classification~\citep{geirhos2018imagenettrained}.

    \subsection{Learning strategies}
    Deploying an appropriate learning strategy in model training is a key to achieving optimal model performance. Techniques such as  contrastive learning~\citep{chen2020simple,khosla2021supervised} and disentangled learning~\citep{du2020learning} enforce models to learn image representations that remain invariant to non-semantic changes in input images. Corruption robustness improves as models learn to ignore the influence of appearance-related changes in images. These strategies implicitly   regularize the models during training. There are also explicit regularization techniques applied to the activation functions, outputs or weights of models. Using multiple models,  where a  model distils knowledge from a  (usually) larger, more complex one, has been shown to benefit model performance.     
    
\begin{figure}[!t]
    \centering
    \includegraphics[width=0.8\linewidth]{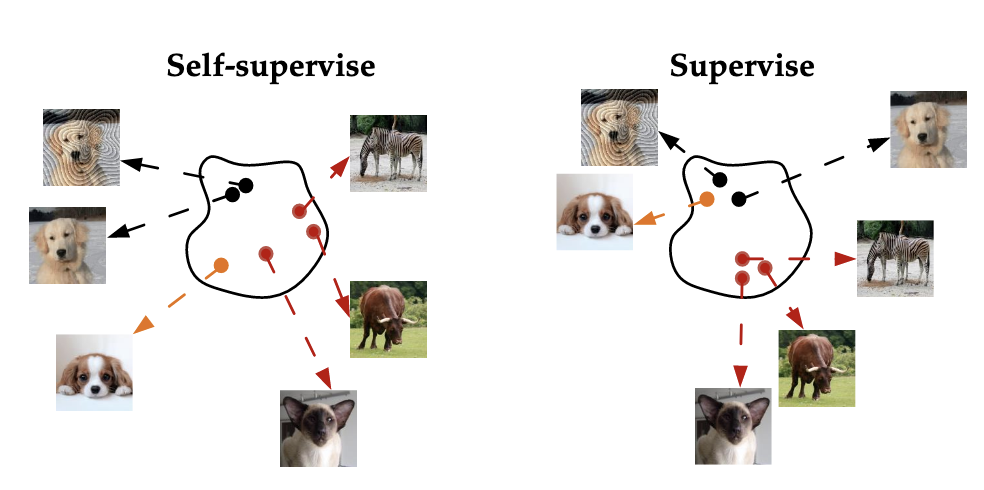}
    \caption{Self-supervised~\citep{chen2020simple} and supervised contrastive learning~\citep{khosla2021supervised}. Without ground truth labels, self-supervised learning may result in learning representations of images from the same class far away from each other in the latent space. For supervised learning, although the image representations of the same class are closer in the latent space, this might result in class collapse when images from the same class point to the same representation.}
    \label{fig:supcon}
\end{figure}

    \textit{Contrastive learning.} 
    Self-supervised learning is shown to be beneficial to improve the robustness of computer vision models~\citep{210412928,190612340,NEURIPS2020_fcbc95cc}. For instance, self-supervised contrastive learning~\citep{chen2020simple} forces a model to learn similar latent representations for source images and their augmented versions, inducing invariance to appearance (non-semantic) changes. Images augmented from the same source image are called positive samples, while the rest are negative samples. The latent representations of the positive pairs are pulled together and those of the negative samples are pushed away from the positive samples (see~\cref{fig:supcon}), e.g. using a perceptual similarity metric~\citep{210606620}. The work in~\citep{NEURIPS2020_ba7e36c4} combines adversarial training with contrastive learning. The problem with these methods is that the latent representations of the same class are also far away from each other, potentially influencing classification tasks, which need class-wise clustered representations. To avoid this,  the authors of~\citep{chakraborty2020gsimclr} additionally use k-means to generate pseudo labels, grouping images with different pseudo labels into the same batch during training.
    
     Supervised contrastive learning (SupCon)~\citep{khosla2021supervised} forces the image representations of the same class to be close to each other. Meanwhile, the representations of other classes are pulled away (see~\cref{fig:supcon}). However, this method trains with a large batch size, increasing computational costs.  SupCon can also  result in class collapse, where all data samples of a class map to the same point in the latent space.  
     Chen~\emph{et al.}~\citep{pmlr-v162-chen22d} proposed a weighted class-conditional loss based on noise-contrastive estimation (InfoNCE~\citep{oord2019representation})  to control the degree of representation spread.   The loss helps better estimate the density of the distribution of positive samples, avoiding class collapse. As contrastive learning might be influenced by spurious correlation, where the learned representations do not completely catch meaningful semantics,~\citep{pmlr-v162-zhang22z} utilizes a model based on  empirical risk minimization (ERM)   to predict spurious attributes, which are further used to prepare positive and negative pairs. Contrastive learning induces invariance to appearance changes. However, its training often requires large batches of images to achieve state-of-the-art performance (e.g. SimCLR~\citep{chen2020simple}), which limits its  research progress and utility in environments with limited computational resources.  

\begin{figure}[!t]
    \centering
    \includegraphics[width=0.8\linewidth]{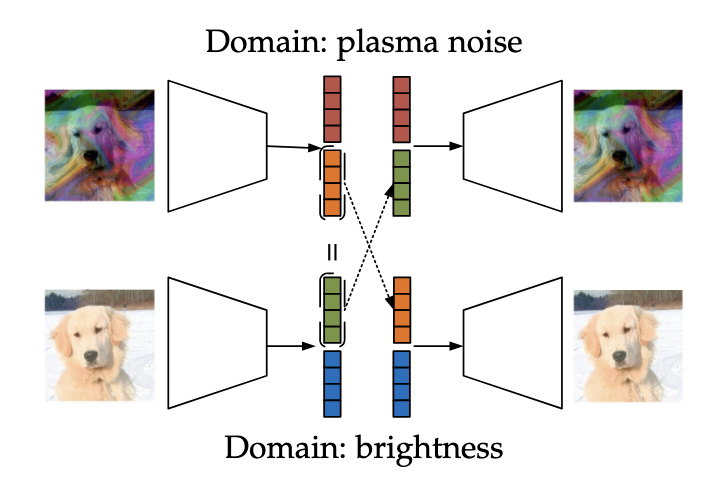}
    \caption{In disentangled learning,  the representation of images is separated into two parts --- style and content codes. The content code represents the semantics while the style code represents appearance and corruption information. }
    \label{fig:Disentanglement}
    \end{figure}

    \textit{Disentangled learning.}
    Disentangled learning enforces invariance to specific characteristics of images by separating the latent content code, which represents the semantics of an image,  and the latent style code, which encodes appearance or other factors related to the operational scenarios. As common corruptions are a form of style, disentangled representations enable models to use  content codes only for tasks like classification. This improves the corruption robustness since the content codes are invariant to different corruptions.
    As illustrated in~\cref{fig:Disentanglement}, style and content codes are learned by cross-domain transfer. A pair of autoencoders is used to transfer images from a domain containing brightness to the other domain containing plasma noise. The content code extracted from the brightness domain together with the style code extracted from the plasma noise domain can be used to generate an image with plasma noise, and vice versa~\citep{du2020learning}.  Being able to extract the content from different styles, models become robust to different corruptions.

    In addition to disentangling representations into style and content encoding,  techniques like invariance-based optimization (IBO) disentangle invariant properties across data collected from multiple environments. Namely, it encourages model to learn invariant representations under different environments~\citep{chang2020invariant,koyama2021invariance}, thus being robust to OOD data. However, current research on IBO usually does not validate their effectiveness in improving corruption robustness, indicating the need for further investigation.

\begin{figure}[!t]
    \centering
    \includegraphics[width=0.8\linewidth]{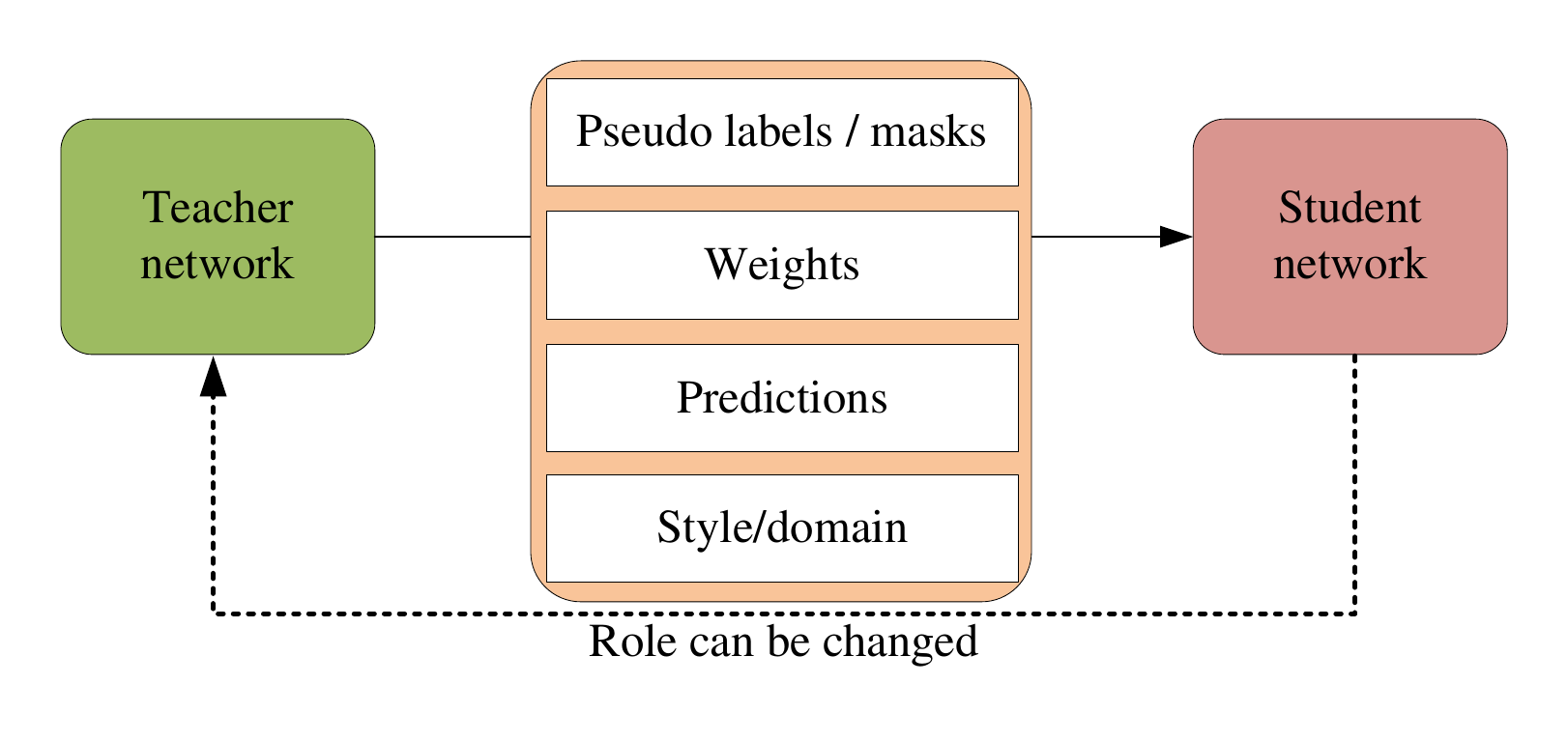}
    \caption{General scheme of knowledge distillation. A student model can distill information from the teacher model, which can be pseudo labels~\citep{9156610}, model weights~\citep{Zhang_2020_CVPR}, segmentation masks~\citep{10019576} model predictions~\citep{Zhang_2020_CVPR} or simulated domain/style information~\citep{anirudh2020mimicgan}. }
    \label{fig:knowledgedistillation}
    \end{figure}

    \textit{Knowledge distillation.}
    Knowledge distillation helps models filter out misguided information and retain meaningful information, thus contributing to increased generalization and corruption robustness. 
    The framework of knowledge distillation needs at least two models: one acts as a teacher, and the other as a student. We provide a general scheme of knowledge distillation in~\cref{fig:knowledgedistillation}. The student network can learn to distill information like pseudo labels / segmentation masks, weights, predictions, and domain information from the teacher network, and their role can be exchanged during training. 
    To improve corruption robustness, the student model is set to distill invariant information w.r.t. image corruption from the teacher model. For instance, the auxiliary classifier in~\citep{Zhang_2020_CVPR} assists the target model to learn invariant image representations regarding image corruptions. The target classifier is trained on original training images while the auxiliary classifier is trained on their augmented versions. Both classifiers share the same backbone, and the auxiliary classifier is trained to predict the same as the target classifier. Thus, the image representations of the original images and their augmented versions are close to each other in the latent space, ignoring the impact of image corruptions.
    
     NoisyStudent~\citep{9156610} trains models by semi-supervised learning. 
     The teacher model is trained on labeled data to provide pseudo labels for the student model, which is trained on a larger unlabelled dataset along with these pseudo labels. The roles of the teacher and student swap in subsequent iterations. The resulting model shows robustness to image corruptions as well as label noise.
     ROCK~\citep{10019576}  applies a segmentation network to create segmentation masks of parts of objects. The segmentation masks of the object parts are compared with prior knowledge about the topology of the object, and the comparison scores will be used for final predictions.  However, this approach requires part labels of objects in a dataset, which are not commonly provided. In knowledge distillation, the performance of the student model largely depends on the quality of the teacher model, and the training is more complex than standard training setups. Nevertheless, knowledge distillation is helpful to reduce the size of the target model while maintaining comparable robustness, making it a promising area for future research.
     
    \textit{Regularization.}
    Regularization can help avoid overfitting and improve the generalization of DNNs. When tested on unknown corruptions, models trained with effective regularization techniques may show better robustness. 
    RoHL \citep{Saikia_2021_ICCV} minimizes the total variations of activations in the first convolutional layer, improving robustness to high-frequency corruptions.  Similarly, JaFR~\citep{ijcai2022p93} enforces the Jacobians of models to have a higher ratio of low-frequency components, improving robustness to high-frequency corruptions. But this introduces a trade-off between robustness to low-frequency and high-frequency corruption. Research in~\citep{Bu_2023_ICCV} shows that controlling the frequency bias of intermediate features can improve adversarial robustness. 
    Regularization-based methods for improving corruption robustness deserve more attention, though being overshadowed by the focus on data augmentation. Proper regularization techniques are based on analyzing model vulnerability, instead of simply feeding models large amounts of data or using large architectures. Moreover, they can often be easily combined with other methods, making them worth further exploration.  

    Commonly used regularization techniques penalize activation functions~\citep{pmlr-v15-glorot11a}, weights of neural networks~\citep{NIPS1991_8eefcfdf,JMLR:v15:srivastava14a,weightuncertain} and output distributions~\citep{pereyra2017regularizing,wang2019learning,lassance2018laplacian,foret2021sharpnessaware}. However, these methods are primarily designed to address overfitting, and have not yet demonstrated their effectiveness in improving corruption robustness, so they are not discussed in this survey.

\subsection{Network components}
    Designing robust network components against data distribution shifts  implies network architectural improvements, such as modifying the receptive field of filters in certain layers of the networks~\citep{krull2019noise2void, laine2019highquality}, normalizing the responses of intermediate layers~\citep{ioffe2015batch, Wu_2018_ECCV,ulyanov2017instance}, and applying attention mechanisms which adjust the learning of models from different parts of inputs~\citep{paul2021vision,Mao_2022_CVPR}. Architectural modifications imply that  complex training strategies~\citep{9156610,chen2020simple,khosla2021supervised,10019576} and large amounts of data~\citep{9156610} are not necessary,  easing computational burden.  
    
    \textit{Receptive fields. }
    Modification of receptive fields  involves changing or extending the kernels of a layer. 
    A network with center blind-spot receptive fields is proposed in~\citep{krull2019noise2void}.  The noise in images is assumed to be pixel-wise independent, and thus the receptive field obtained from the pixels around can help reduce the influence of noise.  
    Combining the receptive fields of a blind-spot network in four directions makes a network intrinsically robust to noise, although not to other corruptions~\citep{laine2019highquality}. Motivated by the biological behavior of neurons in the human visual system, a push-pull layer is proposed in~\citep{RN124} and deployed as a substitute of convolutional layers~\citep{strisciuglio2022visual}. It is based on a response inhibition that suppresses the response to noise and spurious texture, and makes models robust to several unexpected corruptions.  Similarly, the on-/off-kernels~\citep{210607091} are two Gaussian kernels following opposite directions and encourage excitatory center or inhibitory surround, which is shown to be beneficial to robustness under different illumination conditions. BlurPool~\citep{pmlr-v97-zhang19a}, applying low-pass filtering to intermediate feature maps before downsampling them, improves shift-equivariance and robustness. 
    The spatial smoothing kernel~\citep{pmlr-v162-park22b} provides stabilized feature maps that  smooth the loss landscape. This work explains the improved generalization of models using global average pooling, ReLU6, and other similar techniques,  as they have a similar effect on smoothing the responses of models. 
    Recent works on sparse neural networks (SNNs) find that a compact model benefits corruption robustness, compared with its dense version~\citep{winninghand,pmlr-v162-chen22ae}. Instead of training an entire SNN,  Locally Competitive Algorithm (LCANet)~\citep{pmlr-v162-teti22a} uses sparse coding layers in the front end via lateral competition, inspired by a biological observation that neurons inhibit their neighboring neurons which have similar receptive fields.

    \textit{Normalization layers. }
    Normalization layers were originally designed to speed up training, by normalizing the distribution of the inputs of intermediate layers~\citep{ioffe2015batch,ulyanov2017instance, Wu_2018_ECCV}. 
    Subsequently, it was shown that Batch Normalization (BN) layer improves the robustness of models towards image corruptions~\citep{benz2021revisiting,NEURIPS2020_85690f81} and domain shift~\citep{chang2019domainspecific, li2016revisiting}, by simply rectifying the statistics of BN layers according to the new test data. However, BN requires a batch of images to calculate the representative mean and variance. A single image is not enough to obtain representative statistics for the test data, limiting the deployment of BN statistics correction in real applications. 
    Instance normalization (IN) was originally designed to achieve better style transfer, where the feature maps are normalized using the mean and variance of each channel of each sample~\citep{ulyanov2017instance}. It was demonstrated in~\citep{xu2018effectiveness} that IN is robust to specific types of corruption (haze, fog and smoke). Instead of normalizing in the spatial domain, convolutional normalization~\citep{liu2022convolutional} performs normalization in the frequency domain of convolutional kernels, which is shown to be robust to adversarial attacks and label noise. 
 
    \textit{Attention mechanisms.}    
     Transformers have recently gained widespread application in computer vision for their outstanding performance. What sets transformers apart from CNNs is their attention mechanism.
     The work of~\citep{bhojanapalli2021understanding} investigated the robustness of transformers and found that the patch size of Vision Transformer (ViT) is important to robustness, with smaller patches improving adversarial robustness. Furthermore, they found that the robustness of ViTs against common corruptions is comparable to that of ResNets. Another study~\citep{paul2021vision} demonstrated that self-attention mechanism improves the robustness of ViTs against common corruptions, compared to their CNN counterpart, BiT.  Additionally, ViTs were found to be more robust to patch-wise corruptions but more susceptible to adversarial patches compared to CNNs~\citep{gu2022vision}. Replacing  the downsampling blocks in CNNs with the patchifying approach used by ViTs has been shown to benefit corruption robustness of CNNs~\citep{wang2023can}.
     The authors in~\citep{Mao_2022_CVPR} uncovered that classification token (CLS) and locality constraints when calculating pair-wise attention among non-overlapping local patches  impair robustness. Based on these findings, they design a new architecture using robust components like patch/position embedding, called Robust Vision Transformer (RVT). The work in~\citep{pmlr-v162-zhou22m} proposed fully attentional networks (FANs) which add a channel attention layer to improve the visual grouping of mid-level representations.
     Although several studies have shown that attention mechanisms can improve corruption robustness, more investigations are needed to obtain a comprehensive understanding of the capability of self-attention layer to enhance corruption robustness. This is because different works~\citep{bhojanapalli2021understanding,paul2021vision,gu2022vision} reported inconsistent findings.

\subsection{Comparison of methods}

\begin{table}
    \tiny
	\centering         
	\caption{Performance of methods addressing corruption robustness. Results are collected from the papers, including  error rate on ImageNet ($\mathrm{err}$), and robust error rate ($\mathrm{rerr}$) and mean corruption error ({$\mathrm{mCE}$} scaled by  \%)  on ImageNet-$\mathrm{C}$.  Model architecture is ResNet-50. Baseline model is AlexNet.  The best results are in bold, and the best results of each category are underlined. }          
	\label{tab:resultsImageNet}      
     \SetTblrInner{rowsep=1pt} 
	\begin{tblr}{
                hlines,
                rows = { font= \scriptsize},  
                cell{2}{1} = {r=17}{},
                cell{19}{1} = {r=4}{},
                cell{23}{1} = {r=4}{},
                column{1} = {l,0.075\textwidth},
                column{2} = {l,0.2\textwidth},
                column{3-4} = {c,0.02\textwidth},
                column{5} = {c,0.03\textwidth},
                hline{1,2,27} = {-}{1pt},
                hline{3-18} = {-}{0pt},
                hline{20-22} = {-}{0pt},
                hline{24-26} = {-}{0pt},
                }     
	\bfseries Category  & \bfseries Method   & \bfseries $\mathrm{err}$  &\bfseries  $\mathrm{rerr}$   & \bfseries $\mathrm{mCE}$   \\
      
 Data \newline augmentation 
    &   AugMix+SIN~\citep{hendrycks2020augmix}  &25.2 & ---  &   64.9   \\
    &  AutoAug.~\citep{hendrycks2020augmix} & 22.8& ---  & 72.7       \\
    &   RobustMix~\citep{2022robustmix} & 22.9 &  --- &  61.2 \\      
    &    NoisyMix~\citep{NoisyMix} & {22.4} &   {47.7} &  --- \\
    &    PixMix~\citep{Hendrycks_2022_CVPR} & 22.6 &  --- &  65.8   \\
    &    SmoothMix~\citep{Lee_2020_CVPR_Workshops} & 22.34 &  56.64 & ---   \\
    &   PRIME~\citep{prime} & 23 & \underline{45}  &  \underline{57.5}  \\
    &    DDA~\citep{Gao_2023_CVPR} & 23.4 & 70.3  & ---    \\
    &  TestAug~\citep{NEURIPS2020_2ba59664} & 24.1 & 73.61 & ---     \\
    &    GPS~\citep{molchanov2020greedy} &\underline{\textbf{19.07}} &  --- & 67.3     \\
    &    TeSLA~\citep{Tomar_2023_CVPR} & ---& 55  & ---   \\
    &    Cyclic-TTA~\citep{pmlr-v180-chun22a} & 23.81 &  --- &  72.74   \\
    &    SuccessTTA~\citep{10448390} & 24.16 & ---  & 73.51    \\
    &    APR-SP~\citep{Chen_2021_ICCV} & 24.4 & ---  &  65  \\
    &    VIPAug~\citep{Lee_Lee_Myung_2024} & 24.1  & ---  & 65.8    \\
    &    AFA~\citep{vaish2024fourierbasis} & 23.5  & 46.2  & 68     \\
    &    HybridAugment++~\citep{Yucel_2023_ICCV} & 23.7 & ---  &  67.3   \\
    
  Learning  \newline strategies &   SupCon~\citep{khosla2021supervised} & ---  & ---  &   67.2    \\  
    &    REALM~\citep{Seto_2024_WACV} & --- &  \underline{62.2} & ---   \\
    &    MEMO~\citep{NEURIPS2022_fc28053a} &--- & ---  &  69.9   \\
    &    RoHL~\citep{Saikia_2021_ICCV} & \underline{22.7} & ---  & \underline{\textbf{47.9}}   \\
    
    Network  \newline component        
    &   AdaBN~\citep{benz2021revisiting} & --- & 51.3 & 64.9   \\
    &    CorrectBN~\citep{NEURIPS2020_85690f81} & 49.3 &  --- & 62.2   \\
    &    BlurPool~\citep{pmlr-v97-zhang19a} & ---& ---  &  \underline{58.1}   \\
    & P16-Down-Inverted-DW~\citep{wang2023can}   & \underline{22}  &  \underline{\textbf{43.8}} & --- \\
	\end{tblr}
\end{table}

\textit{Computations.}
Techniques applied during inference, e.g. test-time-augmentation~\citep{NEURIPS2020_2ba59664,10417413}, introduce extra computations, thus reducing inference speed. For safety-related scenarios which requires fast responses from systems, this type of techniques might not be suitable. Moreover, test-time augmentation methods reviewed in this survey do not demonstrate significantly superior robustness performance compared to other training-time techniques. Further investigation in test-time augmentation techniques are worth to pursue, for their flexibility in handling new data at inference time. 

Among training-time techniques, contrastive learning and knowledge distillation are the most computationally intensive, either requiring long training time or large-scale datasets. Despite this computational burden, they do not outperform RoHL~\citep{Saikia_2021_ICCV}, a regularization method. This shows that investing large amounts of computational resources does not necessarily bring effective improvements of model robustness. Most training-time data augmentation techniques require comparable computational resources, except for adversary- and learning-based approaches. The effectiveness of methods in these two categories, especially PRIME~\citep{prime},  in improving corruption robustness addresses the importance of designing strategies to make the best use of computational resources.

\textit{Performance.}
We report results collected from the literature of methods addressing corruption robustness on the ImageNet-$\mathrm{C}$ benchmark datasets in~\cref{tab:resultsImageNet}. Robustness analyses of ImageNet-trained models are done solely on ImageNet-$\mathrm{C}$ without considering other benchmark datasets such as ImageNet-$\mathrm{\bar{C}}$ and ImageNet-$\mathrm{3DCC}$. 
     Most methods with benchmark results rely heavily on data augmentation techniques, while many  of them do not follow a standard evaluation process to measure corruption robustness, e.g providing $\mathrm{mCE}$ results. This highlights the need to establish a standardized evaluation procedure to ensure that models are tested consistently, allowing for a fair comparison of performance.

    GPS~\citep{molchanov2020greedy} improves model accuracy on the original ImageNet test set, but its effectiveness at improving corruption robustness is not significant. Instead, from the rerr and $\mathrm{mCE}$ values, P16-Down-Inverted-DW~\citep{wang2023can} and RoHL \citep{Saikia_2021_ICCV} are the two best approaches handling corrupted images, followed by BlurPool~\citep{pmlr-v97-zhang19a}. 
    The effectiveness of these approaches in improving corruption robustness emphasizes the importance of architectural and training strategy improvements in achieving robustness, while larger research interests are mostly focused on developing new data augmentation techniques.
    
    Among data augmentation approaches, PRIME benefits corruption robustness significantly, as evidenced by its low rerr and $\mathrm{mCE}$ values. This effectiveness is attributable to the selection of data augmentations with maximum entropy.  Due to the lack of complete results of some methods,  further comparison among other approaches are challenging. We thus design a unified and standardized corruption robustness evaluation framework in~\cref{sec:benchmark_results}, and compare the performance of popular vision backbones with different model parameters and pre-training data scale.

\begin{table*}
\tiny
	\centering                  
	\caption{Robustness results of backbones trained on ImageNet-1K. The top-3 $\mathrm{mCE}$, $\mathrm{rCE}$ and $\mathrm{RA}$ (scaled by $\%$)  on each benchmark dataset are highlighted in bold and the best results  of CNN and transformer are underlined. Baseline: AlexNet.}
	\label{tab:resultsImageNetbackbone}     
    \SetTblrInner{rowsep=1pt}
    \begin{tblr}{
            hlines, 
            rows = { font= \scriptsize },
            cell{1}{4} = {c=12}{},
            cell{1}{2} = {r=3}{},
            cell{1}{1} = {r=3}{},
            cell{1}{3} = {r=3}{},
            cell{2}{15} = {r=2}{},
            cell{2}{4} = {c=3}{c},
            cell{2}{7} = {c=3}{c},
            cell{2}{10} = {c=3}{c},
            cell{2}{13} = {c=2}{c},
            cell{4}{1} = {r=14}{},
            cell{18}{1} = {r=10}{},
            column{1} = {l,0.07\textwidth},
            column{2} = {l,0.1\textwidth},
            column{3} = {c,0.045\textwidth},
            column{4-13} = {c,0.035\textwidth},
            column{14} = {c,0.04\textwidth},
            column{15} = {c,0.025\textwidth},
            hline{1,4,18,28} = {-}{1pt},
            hline{3} = {4,7,10,13}{leftpos = -1},
            hline{3} = {6,9,12}{rightpos = -1},
            hline{5,6,8,10,11,13,15,16,19,20} = {-}{0pt},
            hline{22,24,26,27} = {-}{0pt}
            }
   \bfseries Architecture &  \bfseries Networks 	& \bfseries  \#Params (M) &\bfseries ImageNet  &   & &  \\
	 &	&   &  $\mathbf{{C}}$ & & & $\mathbf{\bar{C}}$ & &  &  $\mathbf{3DCC}$ & &  & $\mathbf{P}$  & & $\mathbf{SA}\uparrow$\\  
    & &  &   $\mathbf{mCE}\downarrow $  & $\mathbf{rCE}\downarrow $ &$\mathbf{RA}\uparrow$  &  $\mathbf{mCE}\downarrow  $ &  $\mathbf{rCE} \downarrow $ &$\mathbf{RA}\uparrow$ &  $\mathbf{mCE} \downarrow $  &  $\mathbf{rCE}\downarrow  $ &$\mathbf{RA}\uparrow$ & $\mathbf{mFP} \downarrow $ & $\mathbf{mT5D}\downarrow  $  \\
     
     CNN &ResNet-50& 25.56  & 67.98& 98.33& 46.4& 68.74& 98.6& 47.18 & 61.72 & 85.36 & 53.41&48.1 & 95.43 & 80.09\\
    & ResNet-101 & 44.55  & 56.65& 76.88& 55.37&57.62 &76.18 & 55.41 & 54.03 & 72.38 & 59.28& 38.41& 67.38  &81.9\\
    & ResNet-152 & 60.193  & 56.95& 79.54& 55.05& 58.43& 81.41& 54.96 &55.13  &77.2  &58.4 &36.33 & 62.69&\underline{82.55}\\
    &ResNeXt50 & 25.05  &69.75 &99.01 &44.54 & 72.92&106.51 &43.83  &66.4  &94.44  &49.57 & 48.18& 80.22 & 79.66\\
    & ResNeXt101 & 126.89 &  \underline{52.89}& \underline{69.79}& \underline{58.57}& 53.46& 67.77& 58.76 & \underline{52.72} &\underline{70.27}  &\underline{60.39} & 33.95&  58.35 &82.15\\
    &DenseNet121 & 7.98  & 72.05 & 93.17& 43.16& 76.25 &102.51 &41.74  & 68.19 & 85.56 & 48.51& 56.21& 80.95 & 75.57\\
    &DenseNet161 & 28.681  &69.12 & 91.09& 45.38& 72.32& 97.16& 44.5 & 67.88 &90.36  &48.77 & 46.88& 69.52 &77.14\\
    &DenseNet201 & 20.014 & 71.4& 95.52& 43.52& 73.12& 99.01& 44.05 & 69.6 & 93.65 & 47.4&49.2 & 71.54&76.89\\
    &EfficientNet-B2& 9.11   & 66.83& 86.41& 46.95& 65.82& 83.58& 49.46 & 64.39 & 83.51 & 51.31& 39.89& 63.25 &77.78\\
    &EfficientNet-B4& 19.342   &58.87 & 74.38 & 53.7& 56.19& 66.53& 57.16 & 58.77 &75.39  &55.79 &34.32 & \underline{52.78} &79.28\\
    &CoatNet-nano & 15.14  & 60.34& 72.32& 52.66& 60.6& 70.2& 53.75 & 62.61 &79.1  & 52.98& 46.29& 75.84 &77.38\\
    &CoatNet-0 &  27.436  & 60.99& 78.8& 52.14& 58.62& 70.04& 55.16 & 67.41 & 95.43 & 49.2& 39.46& 79.84&78.85\\
    &CoatNet-1 & 41.722  & 54.05& 71.4& 57.62& \underline{50.01}& \underline{59.31}& \underline{61.83} &57.01  &79.43  &57.13 & \underline{30.51} & 80.29&81.69\\
    &ConvNext-T & 28.585  &55.45 &75 & 56.42& 55.86& 73.66&56.76 & 55.64 & 72.05 & 58.4& 31.73& 70.23 &81.87\\
    Transformer     
    &  VOLO-D3& 86.33   &46.92 &63.23 & 62.7& 47.84&62.73 &62.66  &45.86  &\textbf{62.72} & 65.2& \textbf{21.93}& \textbf{48.52}&\textbf{85.23}\\
    &  VOLO-D4& 192.96  &44.77& 61.47& 64.71&46.89 & 63.95& 63.63 & \textbf{45.43} & 64.57 & \textbf{65.76}& \textbf{20.27}& \textbf{48.75}&\textbf{85.87}\\
    &  VOLO-D5& 295.46 &\textbf{42.94} & \textbf{57.34} & \textbf{66.06}& 45.09&59.36 &64.87  & \underline{\textbf{43.35}} & \underline{\textbf{60.08}} & \underline{\textbf{67.29}}& \underline{\textbf{19.45}}& \underline{\textbf{48.03}} &\underline{\textbf{86.07}}\\
     &EdgeNeXt-S& 5.587   &61.19 &81.86 &51.82 &60.73 &77.05 &53.02  &59.05  &78.49  & 55.47& 33.61& 58.67&80.01\\
     & EdgeNeXt-B& 18.51 & 53.16& 72.11& 58.1& 54.11& 71.07& 58.03 & 52.59 & 72.85 & 60.38& 23.64& 50.23 &83\\ 
     & MViTv2-B & 51.47  & \textbf{44.47}&\textbf{56.81}& \textbf{65.27}& \textbf{39.16}& \underline{\textbf{42.09}}& \textbf{70.25} & 46.75 &63.27  &65 & 26.77& 71.29 &84.24\\
     &MViTv2-L& 217.993  &\underline{\textbf{41.91}} &\underline{\textbf{53.51}} & \textbf{\underline{67.24}}& \textbf{37.92}& \textbf{42.4}& \underline{\textbf{71.25}} &\textbf{44.89}  & \textbf{61.31} & \textbf{66.36}& 23.64& 67.64&85.1\\
     & MaxVit-S & 68.928  & 49.07& 67.56& 61.44& 41.45& 48.97& 68.6 & 50.75 & 73.94 & 61.93&27.05 &76.57 &84.43\\
     & MaxVit-B & 119.468 & 46.45& 62.85& 63.51& 39.51& 45.79& 70.11 &49.37  &72.15  & 63& 25.6 & 84.5 &84.86\\
     & MaxVit-L & 211.786 & 45.98& 61.77& 63.84 &\underline{\textbf{36.68}} & \textbf{44.13}& \textbf{70.76} & 48.96 & 71.26 & 63.27& 24.61 & 85.13 &84.94\\
 \end{tblr}
\end{table*}

    \section{Robustness of popular vision backbones}
    \label{sec:benchmark_results}
    Results collected from the literature are not always uniformed with respect to $\mathrm{mCE}$, $\mathrm{mFP}$, and $\mathrm{mT5D}$, complicating the comparison among different works. To address this, we introduce a benchmark framework for evaluating the corruption robustness of relevant pre-trained computer vision backbones. It addresses the limitations of existing robustness tests, which mostly focus on the ImageNet-$\mathrm{C}$ dataset and does not cover broadly the corruption robustness scenarios of a model. 
    Our framework standardizes the evaluation process on a set of benchmarks, including  ImageNet-$\mathrm{C}$, ImageNet-$\mathrm{\bar{C}}$, ImageNet-$\mathrm{P}$, and ImageNet-$\mathrm{3DCC}$.  For computing $\mathrm{mCE}$, $\mathrm{mFP}$ and $\mathrm{mT5D}$, we use the AlexNet trained on ImageNet-1k as the baseline to ensure  fair and comparable results across different datasets and models.  

   \subsection{Vision backbones trained on ImageNet-1k }
    Evaluating model robustness to common corruptions is not usually performed when benchmarking new computer vision backbones.  We conducted a thorough investigation of existing backbones using our framework.
    We report the results in~\cref{tab:resultsImageNetbackbone} and~\cref{tab:resultsImageNetbackboneECE}. 
    The top-3 best mCE and rCE on ImageNet-$\mathrm{C}$, ImageNet-$\mathrm{\bar{C}}$, and ImageNet-$\mathrm{3DCC}$, and the top-3 best $\mathrm{mFP}$ and $\mathrm{mT5D}$ on ImageNet-$\mathrm{P}$,  are highlighted in bold in~\cref{tab:resultsImageNetbackbone}. 
    
\begin{figure*}[!t]
        \centering
         \subfloat[]{\includegraphics[width=\linewidth]{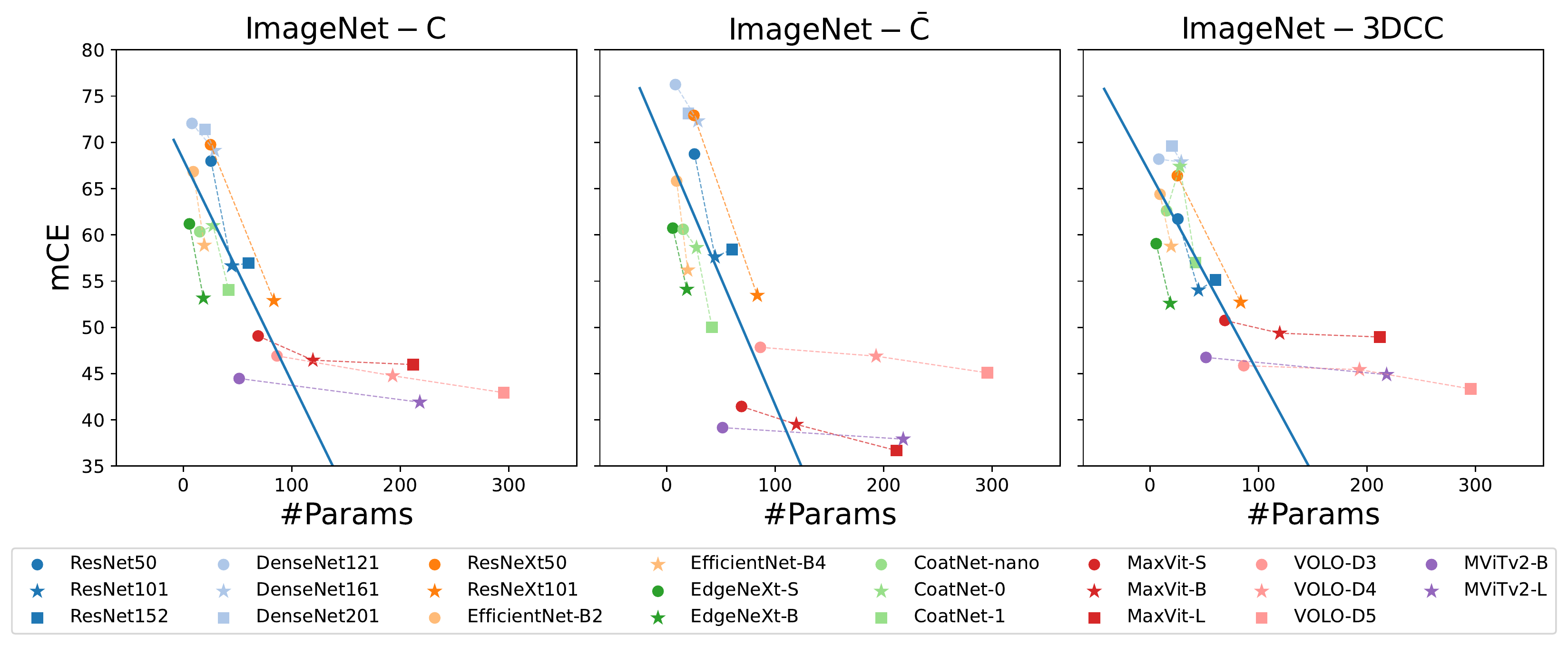}}

         \subfloat[]{\includegraphics[width=\linewidth]{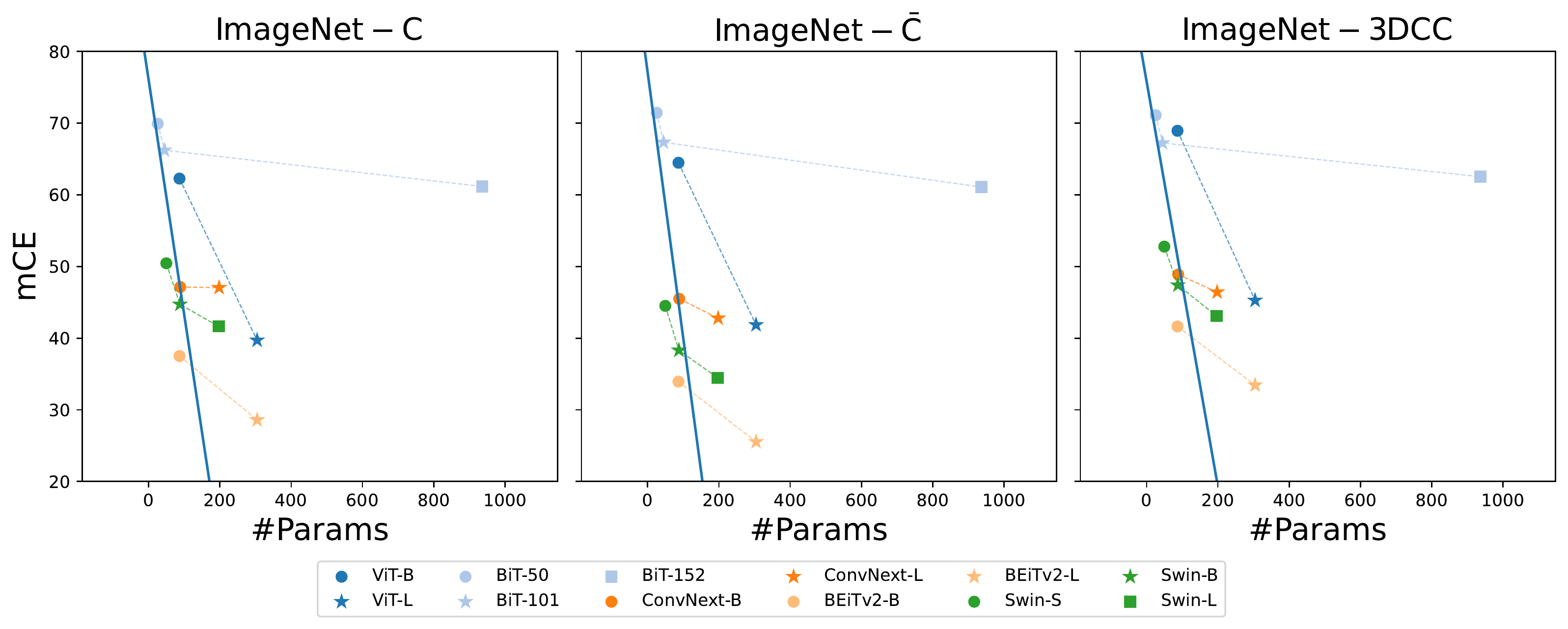}}
        \caption{The size of backbones (a) trained on ImageNet-1k, (b) pretrained on ImageNet-21k and fine-tuned on ImageNet-1k versus $\mathrm{mCE}$ on ImageNet-$\mathrm{C}$, ImageNet-$\mathrm{\bar{C}}$, and ImageNet-$\mathrm{3DCC}$. The mCE does not decrease linearly with an increase in model size. This demonstrates that a larger model is not an effective solution to improving corruption robustness, as the additional computational cost does not provide the expected benefits. }
        \label{fig:plt_imagenet}
    \end{figure*}

     Transformers have shown great potential in improving corruption robustness, as evidenced by the fact that the backbones with top-3 $\mathrm{mCE}$, $\mathrm{mFP}$ and $\mathrm{mT5D}$ are transformer-based. This is not only attributed to their large model size. For instance, even though ResNeXt101 has slightly more parameters than the MaxVit-B transformer, it underperforms MaxVit-B in corruption benchmarks. 
    In~\cref{fig:plt_imagenet}, we analyze the relationship between the number of parameters and the $\mathrm{mCE}$ achieved by 35 backbones on the ImageNet-$\mathrm{C}$, ImageNet-$\mathrm{\bar{C}}$, and ImageNet-$\mathrm{3DCC}$ benchmarks. We observe that, for models with up to about 100 millions parameters, the number of parameters and the $\mathrm{mCE}$ follow a linear relation (illustrated by the blue lines in~\cref{fig:plt_imagenet}). However, increasing the model size beyond a certain point does not necessarily guarantee a corresponding improvement in corruption robustness. The robustness gains for models that have a parameter space larger than 100 millions are indeed very small, despite their much larger complexity and training resources required. 

     Furthermore, we measured the reliability of predictions of these backbones by computing the average normalized $\mathrm{ECE}$ of models tested on ImageNet-$\mathrm{C}$, ImageNet-$\mathrm{\bar{C}}$, and ImageNet-$\mathrm{3DCC}$ in~\cref{tab:resultsImageNetbackboneECE}. 
     The  $\mathrm{ECE}$ of each pre-trained backbone is normalized by that of AlexNet, averaged per corruption type. An  $\mathrm{ECE}$ value smaller than one indicates the predictions of the model are more reliable than those of AlexNet. Many backbones have $\mathrm{ECE}$ values higher than one, indicating less reliable predictions than that of AlexNet, despite better accuracy in robustness tests. 
     For instance, ResNeXt models, with the lowest $\mathrm{mCE}$ among CNNs on ImageNet-$\mathrm{C}$ and ImageNet-$\mathrm{3DCC}$, do not yield the most reliable predictions. Similarly, MViTv2s and VOLOs, achieving better corruption robustness than other transformers,  output less reliable predictions on the corruption benchmark datasets, except for MViTv2-B on ImageNet-$\mathrm{\bar{C}}$.  
    Transformers having superior $\mathrm{mCE}$ (e.g. MViTv2-L and VOLO-D5) do not perform more reliable predictions than many CNNs (they have a much higher average normalized $\mathrm{ECE}$ than many CNNs).  This suggests that progress on vision backbones also requires focus on  addressing robustness and generalization properties explicitly.

     \begin{table}[!t]
     \tiny
	\centering                  
	\caption{Prediction reliability of backbones trained on ImageNet-1k. The average normalized top-3 $\mathrm{ECE}$ on each benchmark dataset are in bold, while the best results of CNN and transformer are underlined. Baseline: AlexNet.}         
	\label{tab:resultsImageNetbackboneECE}     
    \SetTblrInner{rowsep=1pt}
	\begin{tblr}{
            hlines, 
            rows = { font= \scriptsize }, 
            cell{1}{3} = {c=3}{},
            cell{1}{2} = {r=2}{},
            cell{1}{1} = {r=2}{},
             cell{3}{1} = {r=14}{},           
             cell{17}{1} = {r=10}{},
            column{1} = {l,0.1\textwidth},
            column{2} = {l,0.12\textwidth},
            column{3-5} = {c,0.04\textwidth},
            hline{1,3,27} = {-}{1pt},
            hline{4,5,7,9,10,12,14,15,18,19} = {-}{0pt},
            hline{21,23,25,26} = {-}{0pt}
            }     
   \bfseries Architecture &  \bfseries Networks  &\bfseries ImageNet  &   &   \\
	 &	&  $\mathbf{{C}}$ &  $\mathbf{\bar{C}}$ & $\mathbf{3DCC}$ \\  
     CNN &ResNet-50 & 1.137  & 1.514 & 1.588 \\
    & ResNet-101 & \textbf{0.919 }& \underline{1.021} & 1.094   \\
    & ResNet-152 & 0.931 & 1.043 &  \textbf{1.028}\\
    &ResNeXt-50 &1.238 & 1.68 & 1.449  \\
    & ResNeXt-101 & 1.004 & 1.059 & 1.091   \\
    &DenseNet-121 & \underline{\textbf{0.901}} & 1.239 & 1.095   \\
    &DenseNet-161 & 1.401 & 1.447 & \underline{\textbf{0.967}}  \\
    &DenseNet-201 & 1.235 & 1.529 & 1.384 \\
    &EfficientNet-B2& 1.011& 1.056 & 1.233   \\
    &EfficientNet-B4& 1.529 & 1.72 & 1.864 \\
    &CoatNet-nano & 1.06 & 1.092 & 1.304   \\
    &CoatNet-0 & 1.35 & 1.404 & 1.516  \\
    &CoAtNet-1 & 1.523 & 1.194 & 1.831  \\
    &ConvNext-T & 1.346 & 1.343 & 1.7  \\
    Transformer     
    &  VOLO-D3& 1.734 & 1.671 & 1.836 \\
    &  VOLO-D4& 2.274 & 1.907 & 2.387  \\
    &  VOLO-D5& 2.284 & 2.017 & 2.223\\
     &EdgeNeXt-S& \underline{\textbf{0.669}} & 1.142 & 1.149  \\
     & EdgeNeXt-B& 0.935 & 1.112 & 1.081  \\ 
     & MViTv2-B & 1.005 & \textbf{0.957} & 1.122 \\
     &MViTv2-L& 1.454 & 1.282 & 1.668 \\
     & MaxVit-S & 1.041  & \underline{\textbf{0.904}} & 1.083\\
     & MaxVit-B & 1.063 & \textbf{0.957} &  1.157  \\
     & MaxVit-L & 0.977 & \textbf{0.957} & \underline{\textbf{0.983}}  \\
 \end{tblr}
\end{table}

    \begin{table*}
    \tiny
	\centering                  
	\caption{Corruption benchmark results of ImageNet-1K backbones that either distill knowledge from large-scale datasets or are pre-trained on large-scale datasets and  fine-tuned on ImageNet-1K. The best results are in bold and the best results of each network architecture are underlined.  Baseline: AlexNet.}         
	\label{tab:resultsImageNetbackbone_finetuned}     
    \SetTblrInner{rowsep=1pt}
	\begin{tblr}{
            hlines, 
            colsep = 1mm,
            rows = { font= \scriptsize }, 
            cell{1}{5} = {c=12}{},
            cell{2}{5} = {c=3}{},
            cell{2}{8} = {c=3}{},
            cell{2}{11} = {c=3}{},
            cell{2}{14} = {c=2}{},
            cell{2}{16} = {r=2}{},
            cell{1}{2} = {r=3}{},
            cell{1}{1} = {r=3}{},
            cell{1}{3} = {r=3}{},
            cell{1}{4} = {r=3}{},
            cell{5}{1} = {r=22}{},
            cell{5}{2} = {r=12}{},
            cell{17}{2} = {r=3}{},
            cell{20}{2} = {r=3}{},
            cell{23}{2} = {r=2}{},
            cell{25}{2} = {r=2}{},
            column{1} = {l,0.08\textwidth},
            column{2} = {l,0.07\textwidth},
            column{3} = {l,0.11\textwidth},
            column{4} = {c,0.045\textwidth},
            column{5-14} = {c,0.041\textwidth},
            column{15} = {c,0.05\textwidth},
            hline{1,4,11,27} = {-}{1pt},
             hline{3} = {5,8,11,14}{leftpos = -1},
            hline{3} = {7,10,13,15}{rightpos = -1},
            hline{6,7,9,11,12,14,16,18,19,21,22,24,26} = {-}{0pt}
            }    
      &  \bfseries Pre-trained data   & \bfseries Networks 	& \bfseries  Params \newline (M)  &\bfseries ImageNet  &   & &    \\
  	 &	&   &  & $\mathbf{{C}}$ & & & $\mathbf{\bar{C}}$ & &  &  $\mathbf{3DCC}$ & &  & $\mathbf{P}$  & & $\mathbf{SA}\uparrow$\\  
    & &  & &   $\mathbf{mCE}\downarrow $  & $\mathbf{rCE}\downarrow$ &$\mathbf{RA}\uparrow$  &  $\mathbf{mCE}\downarrow $ &  $\mathbf{rCE}\downarrow$ &$\mathbf{RA}\uparrow$ &  $\mathbf{mCE}\downarrow$  &  $\mathbf{rCE}\downarrow $ &$\mathbf{RA}\uparrow$ & $\mathbf{mFP}\downarrow $ & $\mathbf{mT5D}\downarrow $  \\
    Distillation   & IN-21k   & BiT-50 & 25.55  & 52.79   & 71.15  & 58.45   & 50.19 & 62.88 & 61.51 & 54.46     &77.11& 59.02& 33.41& 59.97 &82.82  \\

    Fine-tuning & IN-21k   & BiT-50  & 25.55 & 70.62  & 80.38  &  47.26  & 72.13 &90.62  & 47.22 & 64.67   & 93.24 & 44.53 & 64.48 & 82.47 & 73.63  \\
     & & BiT-101 & 44.541  & 63.5  &  79.72 & 50.05   & 66.49 & 89.19 & 50.29 & 68.54   & 92.39  & 47.3 & 58.1  & \underline{76.21} & 76.01\\ 
    &  & BiT-152 & 936.533   & \underline{58.93}  &  \underline{76.25} &  \underline{53.69}  & \underline{61.43} & \underline{85.09} & \underline{54.18} & \underline{63.96}   & \underline{88.67} & \underline{50.83} & \underline{51.44} & 81.8   & \underline{78.28} \\
    &  & Vit-B/16 & 86.568 & 61.77 & 77.91  &51.35   & 62.77 & 76.46& 51.8& 66.32 & 88.18& 49.75&50.01& 67.2 & 78.02 \\
    &   & Vit-L/16 & 304.205&   \underline{39.71}  & \underline{46.9}  & \underline{68.77}   &\underline{41.86}  &\underline{48.28}  & \underline{67.95} &  \underline{45.27}  & \underline{59.08}  &\underline{65.97} & \underline{26.65} & \underline{47.2} & \underline{84.37} \\
    &   &Swin-S&  49.606    &50.45& 66.72& 60.38& 44.51& 51.66& 66.27 & 52.78 &73.73  &60.34 &33.94 &70.39 &83.05\\
  &   &Swin-B&  87.77 & 44.72& 58.91& 65.01& 38.31& 42.17& 71 & 47.4 & 65.92 &64.37 &30.2 & 73.5  &84.71\\
  &  &Swin-L&  196.53 &  \underline{41.65}& \underline{54.77}& \underline{67.31}& \underline{34.47}& \underline{36.03}& \underline{73.83} &\underline{43.09}  &\underline{59.06}  &\underline{67.63} &\underline{24.19} &\underline{66.18} &\underline{85.83}\\
  &  &BEiTv2-B & 86.53   & 37.51& 44.83& 70.76& 33.94& 36.21& \underline{74.33} & 41.63 & 54.59 &68.79 & 20.65& 47.38& 86.09\\
  &   &BEiTv2-L & 304.43  & \underline{\textbf{28.59}}& \underline{\textbf{29.71}}& \underline{\textbf{77.83}}& \underline{\textbf{25.54}}& \underline{\textbf{20.04}}& 70.75 & \underline{\textbf{33.45}} & \underline{\textbf{40.63}} & \underline{\textbf{75.01}}& \underline{\textbf{14.53}}& \underline{\textbf{38.12}}& \underline{87.41}\\
  &  &ConvNext-B & 88.59  &47.14 & \underline{67.47}&63.17 &45.49 &61.35 & 65.16 & 48.88 & 73.02 & 63.34& 24.94& \underline{65.54} &85.52\\
  &  &ConvNext-L & 197.77 & \underline{47.05}& 69.55& 63.17& \underline{42.79}& \underline{57.7}& \underline{67.27} & \underline{46.43} & \underline{69.69} & \underline{65.22}& \underline{22.16}& 65.79&\underline{86.29}\\
     & Laion2b     & Vit-B/16 & 86.568  &   50.32 &72.32 &  60.25  & 49.37 & 68.9 & 61.98 &50.99 & 75.59& 61.44 & 25.7 & 53.61   & 85.19 \\
     &   & Vit-L/14 & 304.327 & 48.34  & 71.28  & 61.86  & 43.3 & 58.11 & 66.72 & 49.55 & 75.58 & 62.47 & 24.68   & 53.49 &  86.29 \\
     &   & Vit-H/14 & 632.047   &  \underline{39.84} & \underline{53.85}   & \underline{68.56} & \underline{35.54} & \underline{42.4} & \underline{72.7}   & \underline{41.59}  & \underline{59.36} & \underline{68.6} &\underline{19.21}  & \underline{45.13} & \underline{87.23} \\
      & Laion2b \newline + IN12k  & Vit-B/16 & 86.568  & 46.02 & 64.16 &  63.69 & 47.5   & 66.31 &  63.39 & 48.43 & 71.32   & 63.42 &  26.33 & 54.8 & 85.79 \\
    &   & Vit-L/14 & 304.327   &  40.03 & 53.37 & 68.48  &  38.44  & 48.04 & 70.47 & 44.07 & 63.92   & 66.72 & 24.58 & 54.52 & 86.81\\
    &   & Vit-H/14 & 632.047   & \underline{35.78} & \underline{46.92}  & \underline{71.79} & \underline{33.48} & \underline{40.44} & \underline{\textbf{74.35}} & \underline{38.77}  &\underline{55.09}  & \underline{70.78} & \underline{19.79} & \underline{47.83} & \underline{\textbf{87.99}}\\
    & WIT-400M    & Vit-B/16 & 86.568 & 50.69  & 73.3 & 60.04 & 48.91 &  67.93  &  62.42 & 51.8   & 77.67 & 60.89 & 24.29 & 51.68   &85.11\\
     &  & Vit-L/14 & 304.327 & \underline{38.65} & \underline{53.1}  & \underline{69.93}  & \underline{35.17} & \underline{43.62} &  \underline{73.13}  & \underline{41.4}  & \underline{61.13}  & \underline{68.91}  & \underline{17.52} & \underline{46.04}  & \underline{87.61}\\
    & WIT-400M + IN12k  & Vit-B/16 & 86.568  & 46.8 & 65.71  & 63.11 & 48.04 & 67.17 &62.99    & 49.12 & 72.71 & 62.94&   27.34 &   56.14  & 85.63\\
    &  & Vit-L/14 & 304.327   & \underline{35.19} & \underline{45.96} & \underline{72.35} & \underline{33.69} & \underline{41.13} & \underline{74.25} & \underline{39.36}  & \underline{57.13} & \underline{70.44} &\underline{19.9} &\underline{48.85}  & \underline{87.96}\\
 \end{tblr}
\end{table*}

    \subsection{Vision backbones pre-trained on large-scale datasets}
    Fine-tuning models trained on large-scale datasets has been shown to improve the prediction accuracy  of models~\citep{ridnik2021imagenet21k}. Similarly, knowledge distillation improves model performance. 
    We investigate further if fine-tuning or knowledge-distillation contributes to increasing corruption robustness and how the size of pre-training datasets affects robustness performance. We evaluate BiT, ViT, Swin, BEiTv2, and ConvNeXt models with our benchmark framework and present the results in~\cref{tab:resultsImageNetbackbone_finetuned}. The first BiT-50 is trained with a teacher model (BiT-152) that distills information from ImageNet-21k~\citep{ridnik2021imagenet21k}. Other models are  pre-trained on one or two of large-scale datasets (namely ImageNet-21k, Laion2b, WIT-400M and ImageNet-12k) and then fine-tuned on ImageNet-1k.
    
     Compared to knowledge distillation,  fine-tuning models pre-trained on a large-scale dataset do not necessarily gain more corruption robustness. 
    The fine-tuned BiT-50 has higher mCEs than the distilled BiT-50. The fine-tuned BiT-101 and BiT-152 do not outperform the distilled BiT-50 in terms of common corruptions robustness, despite their much larger capacity.
    The fine-tuned ViT-B/16 (86M parameters) and ViT-L/16 (304M parameters) achieve comparable or better robustness than the distilled BiT-50, but at the cost of much higher computational and memory requirements. 
    The effectiveness of knowledge distillation on improving corruption robustness can be observed from the performance of the  BiT-50 which distilled knowledge from a larger BiT-152 model. This demonstrates that proper utilization of large models needs better consideration, as simply increasing model parameters does not effectively gain corruption robustness, if considering the required computational efforts. 

    \begin{figure*}
        \centering
        \includegraphics[width = \linewidth]{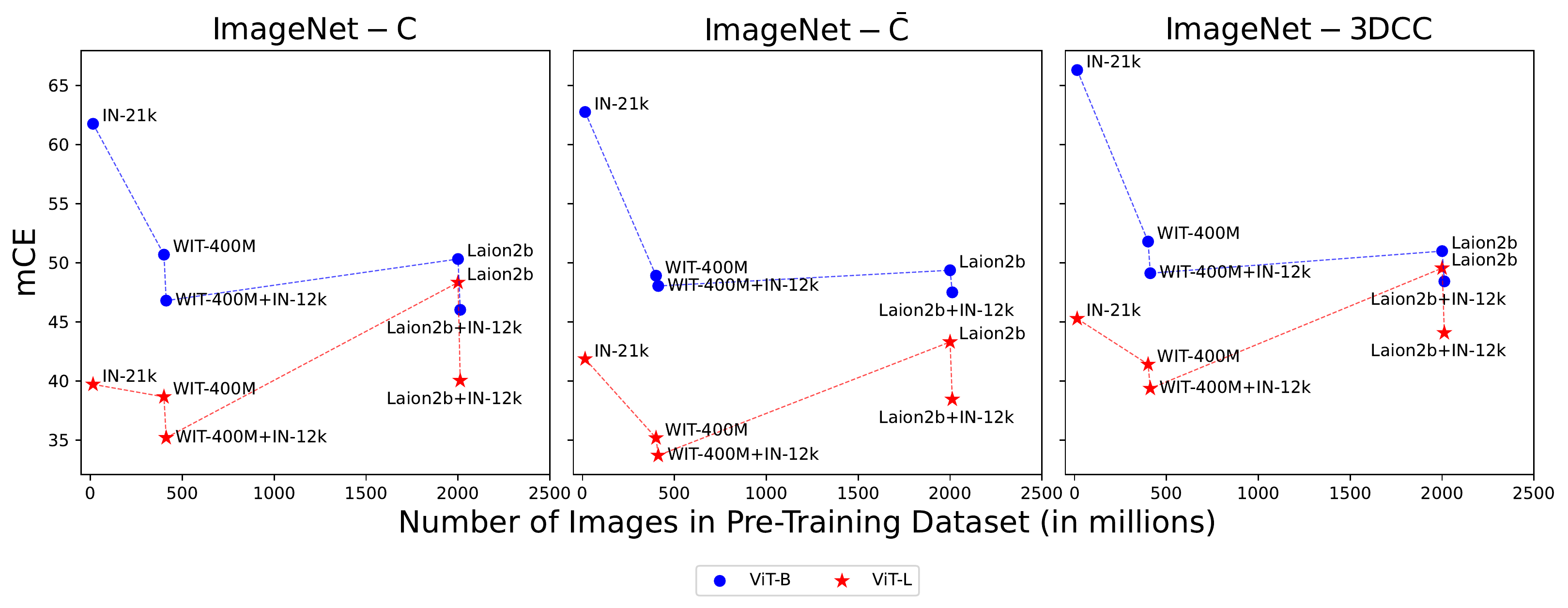}
        \caption{Number of images in pre-training dataset versus $\mathrm{mCE}$ on ImageNet-$\mathrm{C}$, ImageNet-$\mathrm{\bar{C}}$, and ImageNet-$\mathrm{3DCC}$. }
        \label{fig:plot_imagenetbenchmark_dataset}
    \end{figure*}
    We further investigate the impact of pre-training data size on the robustness performance by comparing the results of ViT models in~\cref{fig:plot_imagenetbenchmark_dataset}.  We show that ViT-B achieves the best robustness performance on ImageNet-$\mathrm{C}$, ImageNet-$\mathrm{\bar{C}}$, and ImageNet-$\mathrm{3DCC}$, when pretrained on the largest datasets (Laion2B+ImageNet-12k). ViT-L, instead, achieves the best corruption robustness when pre-trained on WIT-400M+ImageNet-12k, with a comparable performance to ViT-H pre-trained on Laion2b+ImageNet-12k, despite the latter having a much larger model size and using more training data.This indicates that increasing the size of pre-training data, although enabling gains of accuracy on downsteream task benchmarks, does not necessarily account for model robustness performance. 

    Based on these observations, further investigations into techniques focusing on efficient and effective learning for model robustness are required, e.g. methods that efficiently exploit available limited data, leverage prior information, or utilize large models with proper strategies during training. Robustness to image corruptions and generally to visual shifts is thus an open problem that requires an address beyond increasing model and data size.

    \section{Challenges and Future opportunities}
    \label{sec:future}
    Corruption robustness is crucial for  the deployment of computer vision models in safety-critical applications, e.g. autonomous-driving~\citep{7410669}. While there have been significant advances, several challenges and opportunities remain for improving the corruption robustness of these models. We discuss some of these challenges and suggest possible directions for future research.
    
    \textit{Coverage of potential corruption types and levels in the real world.}
     A primary challenge in improving corruption robustness is extensively addressing the diversity of  corruption types and levels encountered in real-world scenarios. They can be more diverse than those included in benchmark datasets, and a model may fail to generalize well to these unexpected corruptions~\citep{taori2020measuring}. Developing methods to handle a wide range of corruptions is critical, especially  given limited data and computational resources. One way is to collect and annotate large-scale datasets that include diverse types of corruption. However, this is time-consuming and expensive. Data augmentation is  currently a research focus, while it needs further investigation to ensure the generated synthetic corruptions are sufficiently representatives of real-world corruptions.

     \textit{Investigation of architectures and  learning strategies.}
     While data augmentation is  effective   for improving corruption robustness, the other two  categories of methods (learning strategies and network components) that focus on structural and architectural changes are promising and need more investigation. 
     Compared to augmentation-based methods which may overfit to corruptions seen during training, there are fewer methods and limited benchmark results for training strategy-based and architectural approaches. Further investigation is required to improve the capability of these robust-by-design methods in learning robust representations that can be transferred to other tasks.
     These techniques are promising for improving model robustness and their potential benefits warrant further research to better understand their capabilities and limitations.

    \textit{Effective and computationally efficient training strategies.}
      While large models and datasets can benefit corruption robustness to some extent, the required computational resources might not always be available. This could limit the reproducibility and usability of the methods. For instance, the backbone of NoisyStudent~\citep{9156610} has over 480M parameters  and was trained using a vast amount of unlabelled images.
      Therefore, it is essential to develop efficient learning strategies that can achieve comparable corruption robustness with limited data and computational resources. Techniques such as transfer learning~\citep{zhuang2020comprehensive}, meta-learning~\citep{hospedales2020metalearning}, and few-shot learning~\citep{song2022comprehensive} can leverage prior knowledge to learn from limited data. 
      Distillation strategy needs to be properly designed, as they show better robustness than fine-tuning models pre-trained on a large-scale dataset.
      Meta-learning uses an outer algorithm to optimize e.g. the learning speed or the generalization performance of the inner algorithm, i.e. the target task. Few-shot learning aims at learning with few or zero samples by matching similar classes that are seen during training.  With these methods,  limited data and computational resources can be used efficiently.

\begin{figure}
    \centering
    \subfloat{
    \includegraphics[width = 0.7\linewidth]{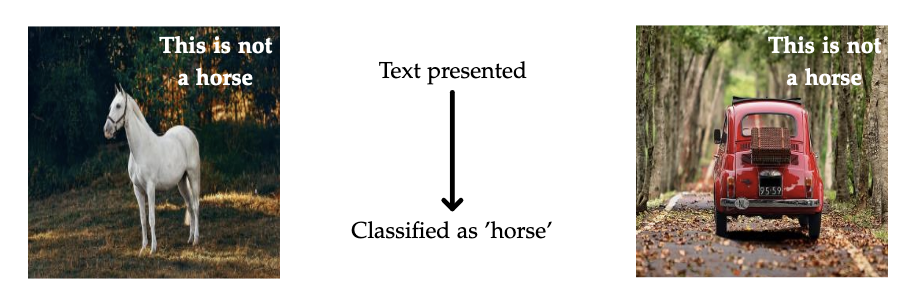}
    \label{fig:text_embeded} }\\
    \caption{Examples of shortcut learning: text embedded in training images of class `horse' forms a spurious correlation between class `horse' and the text, leading to incorrect predictions when the text is in an image of a car~\citep{lapuschkin2019unmasking}.}
\end{figure}

    \textit{Linking corruption robustness, OOD generalization and shortcut learning.}
     Corruption robustness and OOD generalization are closely related, as common corruptions often cause data distribution shifts from the distribution of training data. 
     Shortcut learning, which highlights how models rely on spurious correlations between data and ground truth to achieve high accuracy, also impacts corruption robustness. For instance, in~\cref{fig:text_embeded}, the embedded text in the images of class `horse' induces the models to predict images with the text as a horse, even if  no horses are present~\citep{lapuschkin2019unmasking}. The models do not generalize well to other images with horses but without the text. Such shortcuts  harm the OOD generalization performance of models~\citep{Geirhos_2020}.
     Considering these aspects in the performance evaluation of models can yield more comprehensive and trustworthy results.  Theoretical analysis  that focuses on the learning behavior of models~\citep{shah2020pitfalls,NEURIPS2022_fb64a552} and factors guaranteeing OOD generalization~\citep{xu2022theoretical}  can help better understanding the intrinsic reasons for  impaired corruption robustness, potentially guiding the design of approaches for improving robustness. 
     Developing methods to identify and mitigate shortcut learning (e.g.~\citep{diagnostics12010040,wang2022frequency,wang2023neural,pezeshki2021gradient,ahn2022mitigating,pezeshki2021gradient,06922,06406,08822}) is important to provide a clearer understanding and evaluation of the generalization properties of models, and to enforce the learning of more task-related semantics. Although these methods are not explicitly designed to improve corruption robustness, they may inspire future work on robustness, e.g. one can design approaches to avoid unwanted learning behaviors that impair robustness.

    \textit{Generalization performance of vision-language models.} In benchmark tests,  ViTs trained on large-scale datasets such as Laion2b and WIT-400M (CLIP models) exhibit outstanding robustness performance. However, concerns arise regarding the true generalization capability of these models, particularly when faced with corrupted images. Given that Laion2b and WIT-400M are sourced from the internet, there is a risk of unintentional data leakage. Namely, data (and corruption types) used for evaluation might exist in the training datasets~\citep{chen2024rightwayevaluatinglarge}. Consequently, CLIP models might demonstrate superior robustness performance not because of their inherent generalization ability, but because they have seen similar corruption effects during training. Thus, further research is needed to strictly assess the generalization performance of vision-language models and to ensure a fair comparison with other model types. 
    
    Improving the corruption robustness of computer vision models is challenging and requires developing effective techniques to handle diverse image corruptions. 
    In some cases, applications have hardware constraints at the site of training. Thus, designing robust methods that can operate with limited training data and computational resources is a promising direction to explore. 
    Large-capacity models and large-scale pre-training data are not always feasible due to high computational requirements, though they often achieve good benchmark performance. Smaller models can also be robust if they can distill proper knowledge from other models or datasets. Inducing bias~\citep{20210068} in the form of prior or expert knowledge is a promising research direction. 
    Achieving these goals will foster progress toward the safe and reliable deployment of computer vision models in real-world applications.

    \section{Conclusions}
    \label{sec:conclusion}

    Corruption robustness is a crucial aspect of the performance evaluation of computer vision systems.  It indicates how well a model can operate in the presence of image corruption and provides insights into its effectiveness in challenging environments.
     We surveyed state-of-the-art techniques that address corruption robustness, including data augmentation, learning strategies, and network components. As current benchmark results are not obtained in a unified and coherent way, we constructed a benchmark framework that allows for a unified evaluation of corruption robustness in computer vision. Using this framework, we evaluated popular pre-trained backbones, observing that transformers  exhibit better corruption robustness than CNNs. Large models and pre-training datasets improve corruption robustness. However, the degree of improvement does not  justify the increase in model size and pre-training data, indicating that \textbf{solely scaling up models and pre-training data is not an efficient option to ensure corruption robustness}. Fine-tuning models pre-trained on large-scale datasets does not contribute to more robustness than the model learned with knowledge distillation, demonstrating that \textbf{knowledge transferred from large datasets needs specific design to benefit corruption robustness}.  Based on the benchmark results collected from the literature, strategic and architectural approaches are  currently the most effective for improving corruption robustness, despite most research focusing on data augmentation. 
   
     We suggest focusing on efficient learning techniques that can work with limited computational resources and data, while also exploring the connection among corruption robustness, OOD generalization, and shortcut learning to gain deeper insight into the intrinsic mechanism and generalization abilities of models.

\ifCLASSOPTIONcaptionsoff
  \newpage
\fi

{
\footnotesize
\bibliography{references}
}

\end{document}